\documentclass[lettersize,journal]{IEEEtran}
\usepackage{amsmath,amsfonts}
\usepackage{algorithmic}
\usepackage{algorithm}
\usepackage{array}
\usepackage{textcomp}
\usepackage{hyperref}
\usepackage{stfloats}
\usepackage{booktabs}
\usepackage{url}
\usepackage{verbatim}
\usepackage{multirow}
\usepackage{graphicx}
\usepackage{cite}
\usepackage{arydshln}
\usepackage{overpic}
\usepackage{color}
\usepackage{hyperref}
\usepackage{graphicx}
\usepackage{relsize}
\usepackage{algorithm}
\usepackage{algorithmic}
\newlength\savewidth\newcommand\shline{\noalign{\global\savewidth\arrayrulewidth
  \global\arrayrulewidth 1pt}\hline\noalign{\global\arrayrulewidth\savewidth}}
\newcommand{\tablestyle}[2]{\setlength{\tabcolsep}{#1}\renewcommand{\arraystretch}{#2}\centering\footnotesize}
\renewcommand{\paragraph}[1]{\vspace{1.25mm}\noindent\textbf{#1}}

\usepackage{subcaption, multirow}
\usepackage{bbding}
\usepackage{pifont}
\usepackage{amsmath}

\usepackage{algorithm}
\usepackage{algorithmic}
\newcommand{\app}{\raise.17ex\hbox{$\scriptstyle\sim$}}

\usepackage{xcolor}
\usepackage{newfloat}
\usepackage{listings}
\DeclareCaptionStyle{ruled}{labelfont=normalfont,labelsep=colon,strut=off} 
\floatstyle{ruled}
\newfloat{listing}{tb}{lst}{}
\floatname{listing}{Listing}
\hypersetup{
    colorlinks=true,
    linkcolor=blue,
    filecolor=blue,      
    urlcolor=blue,
    citecolor=blue,
}
\begin{document}

\title{MambaLIE: Scene Light Intensity-Boosted Low-Light Image Enhancement with 
\\
State Space Model}

\author{Wanshu Fan, Xiangyu Li, Cong Wang, 
Kin-man Lam, Xin Yang, Haiyan Zhang and Dongsheng Zhou\

\thanks{This work was supported in part by the National Natural Science
Foundation of China (No. 62502064), Joint Plan of Liaoning Province Science and Technology Plan (No. 2025JH2/101800417), Scientific Research Project of Liaoning Provincial Department of Education
(No. LJ222511258003), Joint Plan of Liaoning Province Science and Technology Plan (No. 2025JH2/101800422), Interdisciplinary Project of Dalian University (No. DLUXK-2025-QN-020), 111 Center (No. D23006). (Corresponding author: Dongsheng Zhou)}
\thanks{Wanshu Fan, Yangyu Li and Dongsheng Zhou are with the National and Local Joint Engineering Laboratory of Computer Aided Design, School of Software Engineering, Dalian University, Dalian, China (E-mail:  fanwanshu@dlu.edu.cn,
lixiangyu@s.dlu.edu.cn
zhouds@dlu.edu.cn).}

\thanks{Cong Wang and Kin-man Lam are with the Hong Kong Polytechnic University, Hong Kong, China (E-mail: supercong94@gmail.com,
kin.man.lam@polyu.edu.hk).}

\thanks{Xin Yang is with School of Computer Science and Technology, Dalian University of Technology
Dalian, China (E-mail: xinyang@dlut.edu.cn).}

\thanks{Haiyan Zhang is with College of Life and Health, Dalian University, China (E-mail: shandongguohua1919@126.com).}

\thanks{The code is available at 
\href{https://github.com/ghfkahfk/MambaLIEcode}{https://github.com/ghfkahfk/MambaLIEcode}.}

}
\date{\footnotesize\textsuperscript{\textbf{1}}National and Local Joint Engineering Laboratory of Computer Aided Design,\\
School of Software Engineering, Dalian University, Dalian 116622, LiaoNing, China\\ }
\markboth{Journal of \LaTeX\ Class Files,~Vol.~14, No.~8, August~2021}%
{Shell \MakeLowercase{\textit{et al.}}: A Sample Article Using IEEEtran.cls for IEEE Journals}


\maketitle

\begin{abstract}
Images captured by consumer electronic devices, such as mobile phones and digital cameras, often suffer from low-light degradation due to sensor limitations and imaging pipelines, which degrades visual quality and affects downstream vision tasks.
Existing methods based on Convolutional Neural Networks (CNNs) and Transformers have dominated current low-light image enhancement (LIE) due to their excellent ability to model hierarchical features.
However, CNNs operate in local receptive fields that cannot model long-range dependencies, while Transformers overcome this problem but incur substantial computational costs.
To address these challenges, we propose MambaLIE, a Scene Light Intensity-Boosted Low-Light Image Enhancement method based on a State Space Model (SSM). 
We first introduce scene light intensity to improve the structural distribution of illumination, which is then gated with the low-light input to guide enhancement.
To better model the illumination while maintaining computational efficiency, we propose the Locally Enhanced State Space Model (LESSM) for efficient light enhancement. 
Our LESSM contains two branches: an SSM branch and a Local Enhanced branch, where the former is used to model the long-range dependencies with linear time complexity, while the latter is used to enhance local feature representations.
Extensive experiments demonstrate that MambaLIE outperforms state-of-the-art CNN-based and Transformer-based LIE methods on four widely used synthetic benchmarks and five publicly available real-world benchmarks in terms of accuracy, speed, and model size, making it suitable for practical deployment on resource-constrained devices.
\end{abstract}

 
\section{Introduction}

\begin{figure}[t] 
\centering
\includegraphics[width=1\linewidth]{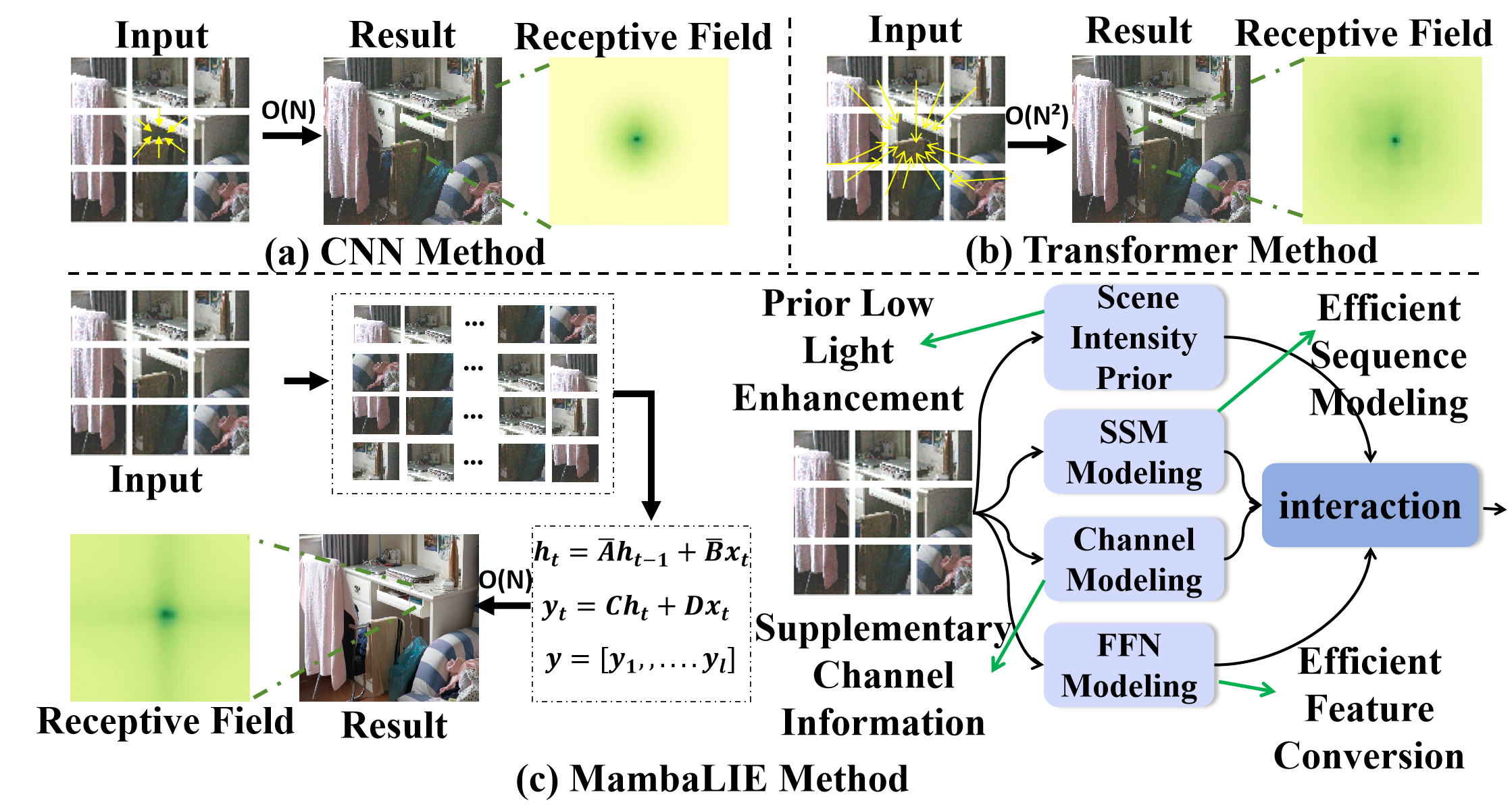}
\caption{Visualization of receptive fields and enhancement results by CNN-based method \cite{cvpr/GuoLGLHKC20}, Transformer-based method \cite{iccv/CaiBLWTZ23} and our proposed MambaLIE. 
In contrast to CNNs that capture only local patterns and Transformers that model global information but introduce quadratic computational complexity with increasing image resolution, our MambaLIE maintains a larger receptive field to achieve global perception with linear computational complexity, thus producing more visually pleasing results.
}
\label{fig:p1}
\vspace{-5mm}
\end{figure}

\IEEEPARstart{L}{ow}-light image enhancement (LIE) plays a critical role in the field of computer vision and image processing, particularly in consumer electronic imaging scenarios such as mobile phones and digital cameras. 
It involves improving the quality and visual appearance of images captured under low light or dim conditions through image processing techniques.
Because of insufficient light and sensor limitations, images often suffer from poor global visibility and local problems, such as color distortion and noise. 
These degraded images can adversely impact human perception and hinder subsequent processing and analysis.
Such degradation is commonly encountered in practical imaging systems~\cite{image,camera}.
Hence, LIE not only contributes to better visual perception, but also improves the efficiency of downstream applications, such as mobile photography and intelligent surveillance, object detection~\cite{zou2023object} and semantic segmentation~\cite{li2023mseg3d}.
LIE is a typical low-level vision task, along with others such as image super-resolution~\cite{chaofen1,chaofen2}, defogging~\cite{quwu1}, and underwater image enhancement~\cite{shuixia1}.
%

%
Early techniques, such as histogram equalization \cite{dsp/ChengS04} and gamma correction \cite{displays/WangLL09}, enhance images through global intensity mapping, but often struggle with complex lighting conditions and preserving naturalness. Low-light imaging is particularly critical for safety-sensitive applications, such as pantograph anomaly detection in tunnels.
Recently, CNN and Transformer-based methods have achieved remarkable performance in this field~\cite{cvpr/GuoLGLHKC20,aaai/WangZSLSL23,cviu/DangZQ24,iccv/CaiBLWTZ23}. However, CNNs are inherently limited by their local receptive fields and weight sharing, making them less effective in modeling globally inconsistent illumination. 
In contrast, Transformer-based approaches capture long-range dependencies via self-attention, enabling better structural and detail recovery. Nevertheless, their quadratic computational complexity leads to high resource consumption and limits practical deployment.

Recently, state space models (SSMs), especially Mamba~\cite{corr/abs-2312-00752,vmamba}, a special mode of SSMs, have attracted significant attention in the field of computer vision. 
Several vision-oriented extensions, including S4ND~\cite{s4nd}, VMamba~\cite{vmamba}, and Vim~\cite{vim}, further demonstrate the effectiveness of state-space modeling for capturing global contextual information.
These internal SSMs show great potential for modeling global information with linear complexity while providing a larger receptive field compared to CNN-based and Transformer-based approaches.
They simplify processing by using selective SSMs to effectively capture important features.
%
Moreover, SSMs scale linearly across sequence lengths, leveraging GPU streams for highly parallelized computations, making them an efficient choice for tasks requiring both speed and scalability.
However, these methods are primarily designed as general-purpose visual backbones and are mainly evaluated on high-level vision tasks such as image classification, object detection, and semantic segmentation.
In contrast, low-light image enhancement is a low-level image restoration problem that requires not only global context modeling but also explicit handling of non-uniform illumination and preservation of fine-grained local structures.
As a result, directly applying existing vision-SSM models to LIE is non-trivial, as they do not explicitly account for illumination variation or local detail restoration.
Hence, it is of great interest to explore how to develop a suitable Mamba into the Low-light Image Enhancement (LIE) task.

\begin{figure*}[t] 
\centering
\includegraphics[width=1\linewidth]{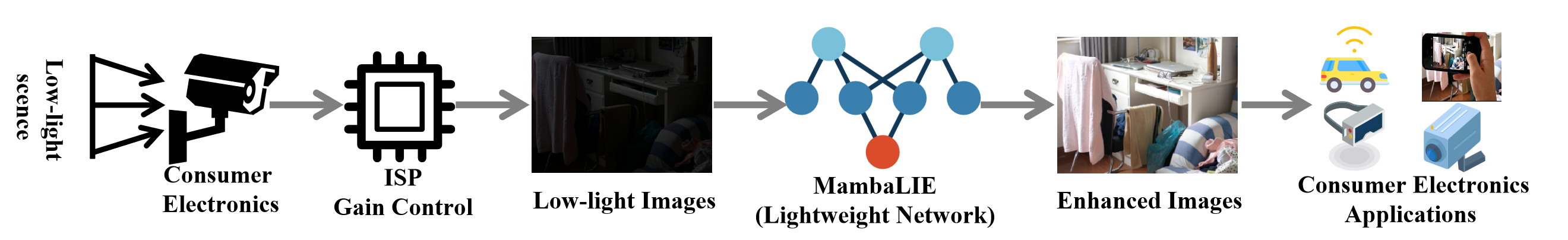}
\vspace{-6mm}
\caption{Overview of the MambaLIE-based low-light image enhancement pipeline in consumer electronic imaging systems.}
\label{fig:p01}
\vspace{-2mm}
\end{figure*}

In this paper, we propose MambaLIE, a state space model boosted with scene light intensity, to solve the enhancement problem in low-light scenarios.
We first introduce a scene light intensity prior to effectively perceive light intensity to help enhance images.
This prior is further gated with the original low-light input image, which serves as the input of the LIE network to reveal more structural content to facilitate enhancement.
Furthermore, we introduce a locally enhanced state space model (LESSM) to effectively and efficiently enhance the gated images.
Our LESSM contains an SSM branch and a local branch.
The SSM branch consists of a forward SSM and a backward SSM, the former processes the input sequence in a standard, chronological order, while the latter processes the sequence in reverse order to help in capturing dependencies that might be missed by the forward pass alone.
The local branch is used to preserve useful local information to further improve feature representations.
By combining LESSM with channel attention~\cite{cvpr/HuSS18} and a feed forward network~\cite{nips/VaswaniSPUJGKP17} within a U-shaped encoder-decoder structure, our MambaLIE achieves excellent enhancement performance while maintaining relatively friendly computational complexity.

Fig.~\ref{fig:p1} provides a visual comparison of receptive field coverage and enhancement results by CNN-based method~\cite{cvpr/GuoLGLHKC20}, Transformer-based method~\cite{iccv/CaiBLWTZ23}, and our proposed MambaLIE.
The comparison shows that MambaLIE produces a larger
receptive field and obtains superior results.
%
%
%

Compared to Transformers, MambaLIE is designed to address the challenge of balancing computational efficiency with enhancement quality in low-light image enhancement. 
Transformer-based methods such as Retinexformer~\cite{iccv/CaiBLWTZ23} and LLFormer~\cite{aaai/WangZSLSL23} adopt multi-head self-attention (MHSA) modules to capture long-range dependencies, but incur quadratic computational complexity with respect to input resolution. 
In contrast, MambaLIE replaces the MHSA backbone with a state-space modeling (SSM) framework, which enable global sequence modeling with linear complexity. Additionally, it introduces a learnable scene light prior through spatial gating, offering a fully data-driven alternative to fixed prior designs. 
This architecture allows MambaLIE to efficiently scale to high-resolution inputs while preserving both fine-grained local textures and global illumination consistency. 
These differences in modeling strategy and structural design distinguish MambaLIE from prior Transformer-based approaches, achieving a more favorable trade-off between performance and efficiency in real-world low-light enhancement.  
%
Thanks to its efficiency and lightweight design, the proposed MambaLIE can be directly applied in consumer electronic devices, such as mobile phones and digital cameras, for practical low-light imaging, as illustrated in Fig.~\ref{fig:p01}.

Our contributions are summarised as follows:
\begin{itemize}
\item We propose MambaLIE, a scene light intensity-boosted LIE with State Space Model, to effectively explore Mamba for modeling global dependencies with linear computational complexity to solve LIE.

\item We propose a scene light intensity prior to capture light information to reveal useful structural content for boosting enhancement quality.

\item Extensive experiments show that our MambaLIE outperforms existing CNN-based and Transformer-based LIE methods while achieving a better trade-off between model complexity and performance on both four widely used synthetic benchmarks and five real-world datasets.
\end{itemize}
\section{Related Work}
In this section, we review related works on low-light image enhancement and vision state space model.
\subsection{Low-Light Image Enhancement}
In the early stages of low-light image enhancement (LIE), methods primarily focused on adjusting pixel intensity distributions to emphasize low-value regions and reveal hidden details. Typical approaches include histogram equalization for contrast enhancement \cite{dsp/ChengS04, tip/LeeLK13}, gamma correction with S-curve transformations to amplify dark regions \cite{displays/WangLL09}, and Retinex-based methods that decompose images into reflectance and illumination components to improve brightness while preserving color consistency \cite{ejivp/RahmanRAAS16, tip/WangZHL13, TciRIRO}.

However, these methods lack semantic awareness and often introduce artifacts such as color distortions and unnatural textures, limiting visual quality.

With the success of convolutional neural networks (CNNs), LIE has achieved significant progress. CNN-based methods enable end-to-end learning of complex features, allowing more effective contextual and structural enhancement than traditional approaches \cite{pwgcm, sclm, tciLFIENet}, and have become the dominant paradigm.

More recently, Transformer-based models have further advanced LIE by capturing long-range dependencies via self-attention mechanisms \cite{cvpr/XuWFJ22, iccv/CaiBLWTZ23, TCIDual-Stream}. Despite their strong performance, the quadratic complexity of self-attention with respect to sequence length leads to high computational cost, limiting their practicality for long-sequence processing.

This computational complexity limits their practicality for certain applications, especially in resource-constrained environments, and underscores the need for models that balance accuracy with efficiency in LIE tasks, particularly in consumer electronic imaging scenarios where real-time processing and efficiency are critical.

\subsection{Vision State Space Model}

The State Space Model (SSM) has recently emerged as a promising framework for sequence modeling in deep learning. Initially applied in natural language processing, SSMs have demonstrated strong capability in capturing long-range dependencies through structured formulations \cite{nips/GuG0R22, iclr/GuGR22}, providing an alternative to CNNs and Transformer-based models.

A key advantage of modern SSMs, such as Mamba \cite{corr/abs-2312-00752}, lies in their ability to model long-range dependencies with linear complexity with respect to input size, addressing the quadratic complexity limitation of Transformers. This efficiency in both computation and memory makes SSMs particularly suitable for long-sequence modeling and has attracted increasing attention across domains.

More recently, SSMs have been extended to computer vision tasks. S4ND \cite{s4nd} incorporates state-space mechanisms into vision architectures to capture spatial dependencies, while Vmamba \cite{vmamba} further adapts SSMs to non-causal visual data by modeling global spatial context. Additionally, Vim \cite{vim} introduces a bidirectional state-space structure with location awareness, enabling more effective feature representation for visual understanding.

Unlike prior vision state-space models, which are primarily developed as general-purpose backbones for sequence or visual representation learning~\cite{nips/GuG0R22, iclr/GuGR22, corr/abs-2312-00752} and later extended to vision tasks~\cite{s4nd, vmamba, vim}, the proposed LESSM is specifically tailored for low-light image enhancement. 
Existing SSM-based vision methods mainly emphasize global sequence modeling, while lacking explicit mechanisms to handle non-uniform illumination and local detail degradation. 
Moreover, these methods are mainly evaluated on high-level vision tasks such as classification, detection, and segmentation, and have not been specifically designed or benchmarked for low-light image enhancement.
In contrast, LESSM introduces a hybrid global–local design together with bidirectional state-space modeling, enabling both long-range dependency modeling and fine-grained structure preservation under spatially varying illumination. 
Furthermore, LESSM incorporates a scene light intensity prior to guide feature transformation, introducing illumination-aware modulation into the modeling process.

Building on these observations, we leverage Mamba's efficient long-range modeling capability for low-light image enhancement (LIE).
Specifically, we use Mamba’s capacity for linear complexity analysis to integrate features informed by retinal theory, aiming to enhance image quality in low-light scenarios. 
The improvements observed in low-light image quality underscore the potential of SSM-based models, like Mamba, to address challenging visual tasks, further validating their role in computer vision applications.
These properties make state-space models particularly suitable for low-level vision tasks under computational constraints.
\begin{figure*}[!t] 
\includegraphics[width=\linewidth]{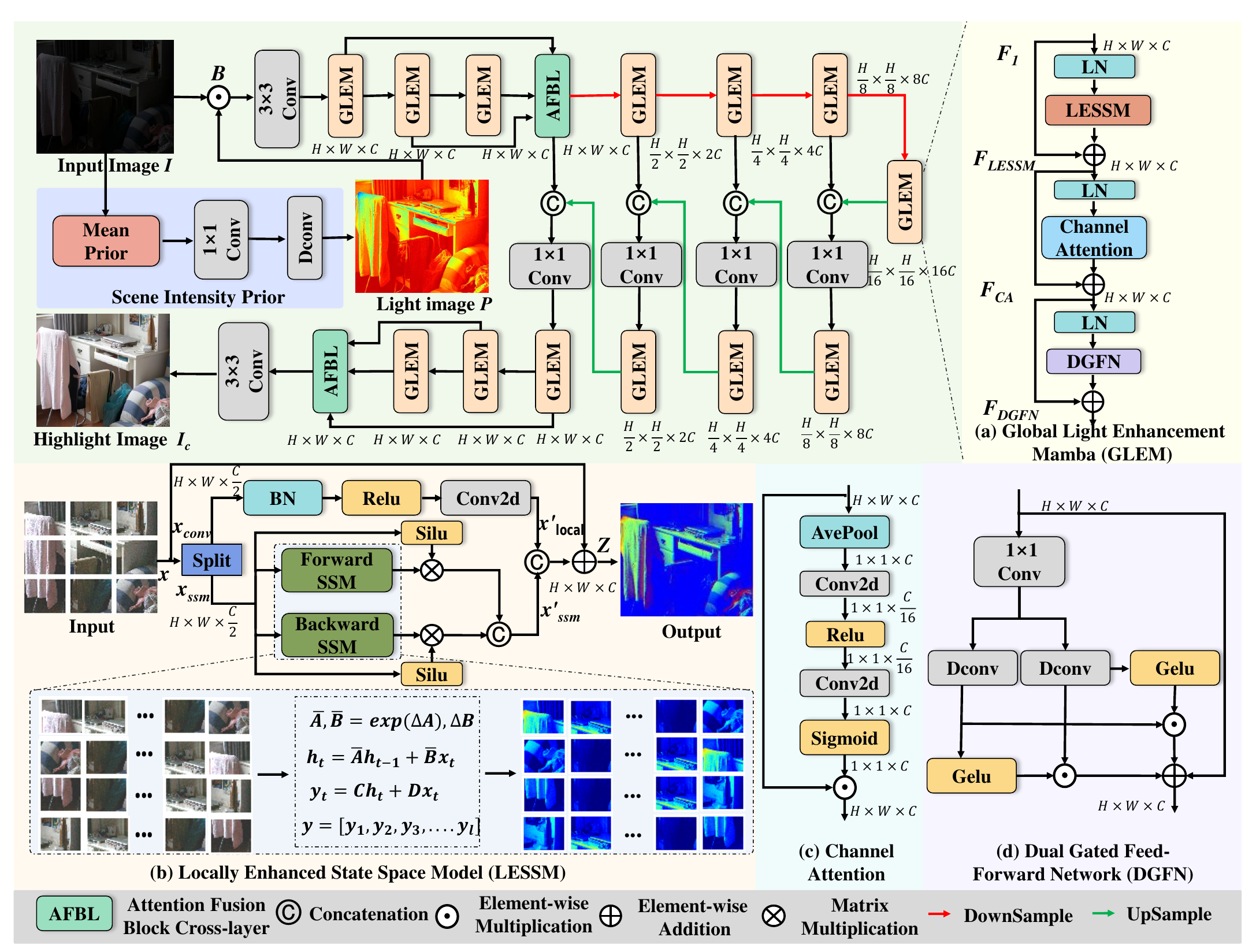}
\caption{The overall architecture of our proposed MambaLIE. 
Our MambaLIE first employs a mean filter to obtain the scene light intensity prior, which produces the structural information of the low-light input image. 
Then, this scene light intensity prior is convolved by several convolutional layers and gated with the low-light input image, 
serving as the input of the network.
The gated image is further enhanced by combining the Locally Enhanced State Space Model (LESSM), channel attention, and feed-forward network within a U-shaped encoder-decoder framework to obtain the final enhancement result.
Our LESSM contains two branches: SSM branch and Local Enhanced branch, where the former is used to learn global information while the latter is used to enhance local feature representations.
}
\label{fig:p2}
\vspace{-2mm}
\end{figure*}

\section{Proposed Approach}

Our goal is to efficiently transform low-light images into normal-light images. 
To this end, we propose MambaLIE, a scene light intensity-boosted low-light image enhancement method utilizing a state space model. 
Our MambaLIE employs a mean filter to obtain scene light intensity, revealing more useful structural information, which is then fused with the original low-light input in the network to boost enhancement. 
To reduce computational complexity and perceive a larger receptive field for better enhancement, we propose the Global Light Enhancement Mamba (GLEM). 
Different from prior vision state-space models that mainly serve as general-purpose backbones~\cite{s4nd, vmamba, vim}, the proposed design is tailored for low-light image enhancement by explicitly incorporating illumination-aware modeling and joint global--local feature learning.
GLEM utilizes the State Space Module (SSM) to perceive global light within linear computation efficiency, and uses channel attention~\cite{eccv/ZhangLLWZF18} to reduce redundant channels, while also incorporating a Dual Gated Feed-Forward Network (DGFN)~\cite{aaai/WangZSLSL23} for efficient feature conversion. 
The overall framework is designed with computational efficiency in mind, making it suitable for practical deployment under resource-constrained environments.

\subsection{Overall Pipeline}
Fig.~\ref{fig:p2} shows the overall pipeline of our proposed MambaLIE, which features a U-shaped encoder-decoder framework with scene light intensity-boosted operation and state space model as the basic block.

Given a low light image $I \in \mathbb{R}^{H\times W\times3}$, MambaLIE first uses a mean filter and convolutions to obtain the scene intensity prior map $P \in \mathbb{R}^{H\times W\times3}$. 
Then, $I$ is gated with $P$ by element-wise multiplication to obtain the boosted result $B$.
MambaLIE employs a 3$\times$3 convolution on $B$ to extract shallow feature $F_{b}\in\mathbb{R}^{H\times W\times C}$.
Next, $F_{b}$ is fed into the Global Light Enhancement Mamba, which contains the state space model, channel attention~\cite{eccv/ZhangLLWZF18}, dual gated feed-forward network~\cite{aaai/WangZSLSL23}, and an Attention Fusion Block Cross-layer~\cite{aaai/WangZSLSL23} to extract deeper features. 
After that, the feature $F_{d}$ is obtained by deep feature extraction on $F_{b}$ in the encoder stage.
%
%
%
Subsequently, the low-resolution latent feature $F_{d}$ is processed in the decoder stage, which progressively reconstructs the high-resolution representations.
%
%
%
%
We employ a weighted skip connection combined with a convolution for feature fusion between the encoder and decoder.
%
%
After the decoder, the deep feature $F_{e}$ is obtained, and $F_{e}$ is then processed by GLEM and the Attention Fusion Block Cross layer~\cite{aaai/WangZSLSL23}.

Finally, MambaLIE applies a 3$\times$3 convolution on the enhanced features to yield the enhanced images $I_{e}$. 
The final reconstruction result is obtained by the following formula:
\begin{equation}
I_{c}=\mathcal{N}(I),
\label{con:1}
\end{equation}
where $\mathcal{N(\cdot)}$ represents the entire network, which is trained by minimizing the following loss function:
\begin{equation}
{L}=\|I_{c}-I_{gt}\|_{1},
\label{con:2}
\end{equation}
where $I_{{gt}}$ denotes the ground truth image and $\|{\cdot}\|_{1}$ represents the $L_{1}$ loss.

\subsection{Scene Light Intensity Prior}

Employing scene light intensity is critical for low-light image enhancement (LIE), as non-uniform illumination often causes the disappearance of structural details and uneven exposure. 
Providing explicit illumination guidance helps the network distinguish between severely underexposed areas and actual image content, which is essential for avoiding over-enhancement and preserving detail.

To this end, we introduce a Scene Light Intensity Prior $P$, which provides a spatially smoothed representation of the scene's global illumination. 
It is computed from the original low-light image $I$ using a mean filter $\mathcal{M}(\cdot)$:
\begin{equation}
P = \mathcal{M}(I).
\end{equation}

This prior retains the global luminance trend while suppressing high-frequency noise, serving as a low-frequency approximation of the scene's illumination distribution.
We then apply a $3 \times 3$ depthwise convolution and a $1 \times 1$ point-wise convolution to $P$ to extract modulation features. These features are used to gate the original input $I$ via element-wise multiplication:
\begin{equation}
B = W_d W_p(P) \odot I,
\end{equation}
where $\odot$ denotes element-wise multiplication, and $W_d$, $W_p$ denote depthwise and point-wise convolutional operators, respectively.


%
\subsection{Global Light Enhancement Mamba}
%
Instead of using convolution that has a limited receptive field, and Transformers that can model long-range dependencies but introduce higher computational complexity, we propose the Global Light Enhancement Mamba (GLEM).
This approach is designed to perceive global light while maintaining linear computational complexity.
%
As illustrated in Fig.~\ref{fig:p2}(a), given the input feature $F_{1}$, we first apply Layer Normalization~\cite{corr/BaKH16} followed by the Locally Enhanced State Space Model to capture global spatial information, as follows:
\begin{equation}
\textit{F}_{\textit{LESSM}}=\textit{LESSM}(\textit{LN}(F_{1}))+F_{1},
\label{con:3}
\end{equation}
where $\textit{LN}(\cdot)$ indicates layer normalization \cite{corr/BaKH16}, and $\textit{LESSM}(\cdot)$ denotes the Locally Enhanced State Space Model, as shown in Fig.~\ref{fig:p2}(b).
%
%

To improve channel modeling ability and minimize channel information redundancy, we introduce channel attention~\cite{eccv/ZhangLLWZF18}, expressed as follows:
\begin{equation}
\textit{F}_{\textit{CA}}=\textit{CA}\left(\textit{LN}\left(\textit{F}_{\textit{LESSM}}\right)\right)+\textit{F}_{\textit{LESSM}},
\label{con:4}
\end{equation}
where $\textit{CA}(\cdot)$ denotes the Channel Attention~\cite{eccv/ZhangLLWZF18}, as shown in Fig.~\ref{fig:p2}(c). 
This integration enables GLEM to focus on learning diverse channel representations, thereby improving channel modeling capabilities and avoiding redundancy.

%
Similar to the pairing of attention and FFN, here we introduce the Dual Gated Feed-Forward Network (DGFN)~\cite{aaai/WangZSLSL23} to enhance feature transformation:
\begin{equation}
\textit{F}_{\textit{DGFN}}=\textit{DGFN}(\textit{LN}(F_{CA}))+\textit{F}_{\textit{CA}},
\label{con:5}
\end{equation}
where $\textit{DGFN}(\cdot)$ denotes the Dual Gated Feed-Forward Network, as shown in Fig.~\ref{fig:p2}(d).
DGFN enhances the model's capability to model non-linear relationships, which is particularly beneficial for handling complex data and optimizing feature representation, thereby enhancing performance and generalization on complex scenes. 
%
%
%

%
\noindent \textbf{Locally Enhanced State Space Model.}
The CNN-based approaches~\cite{mm/ZhangZG19,bmvc/WeiWY018} typically possess a limited receptive field, which restricts model's ability for image restoration or enhancement. 
In contrast, Transformer-based models can provide a larger receptive field but incur significant computational complexity due to global attention mechanisms. 
To address these challenges, we propose a Locally Enhanced State Space Model (LESSM), which can efficiently capture long-range dependencies and model complex visual dynamics while maintaining linear computational complexity.
LESSM contains two branches: the SSM branch and the Local Enhanced branch. The former is used to learn global information, while the latter is used to enhance local feature representations.
The architecture of LESSM is illustrated in Fig.~\ref{fig:p2}(b).
First, the original input $x$ is split into two signals $x_{local}$ and $x_{ssm}$:
\begin{equation}
x_{local}, x_{ssm}=\textit{Split}(x).
\end{equation}
%

%
Next, these two inputs are then fed into the Local Enhanced branch and SSM branch, respectively. 
In the Local Enhanced branch, we use convolution and activation functions to extract local features and enhance feature representation:
\begin{equation}
x'_{local}=\textit{Conv}(\textit{Relu}(\textit{BN}(x_{local}))),
\end{equation}
where $\textit{BN}(\cdot)$ denotes batch normalization.

In the SSM branch, $x_{ssm}$ is divided into four branches.
Two of these branches capture global features through the ForwardSSM and the BackwardSSM respectively, while the remaining two branches pass through the SiLU activation function.
The branches of ForwardSSM and BackwardSSM are multiplied by SiLU and then concatenated.
The operations in the SSM branch can be represented as follows:
%
\begin{equation}
x_{1}, x_{2}, x_{3}, x_{4}=\textit{Split}(x_{ssm}),
\end{equation}
%
%
\begin{equation}
\begin{aligned}
x'_{ssm} = \mathcal{C}[\textit{FSSM}(x_{1})\otimes\textit{SiLU}(x_{2}), \textit{BSSM}(x_{3})\otimes\textit{SiLU}(x_{4})],
\end{aligned}
\end{equation}
where $\textit{FSSM}(\cdot)$ denotes the ForwardSSM, 
$\textit{BSSM}(\cdot)$ denotes the BorwardSSM,
$\otimes$ is the matrix multiplication,
and $\mathcal{C}[\cdot]$ means the concatenation operation.

ForwardSSM and BackardSSM play a crucial role in converting one-dimensional inputs to outputs by employing latent states within a framework based on linear ordinary differential equations.
For the system with input $x_{t}$ and output $y_{t}$, the model
dynamics are described as follows:
\begin{equation}h_{t}=\bar{A}h_{t-1}+\bar{B}x_{t},\label{con:7}\end{equation}
\begin{equation}y_{t}=Ch_{t}+Dx_{t},\label{con:8}\end{equation}
\begin{equation}y=[y_1, y_2,\dots, y_L],\label{con:9}\end{equation}
where $\bar{A}$, $\bar{B}$, $C$ and $D$ denote the model parameters. 
Subsequently, the outputs from the extracted features in the four scanning directions, $y_{1}$, $y_{2}$, $\dots$, $y_{L}$, are combined through summation. 
The merged output is then reshaped to align with the original input size.

Finally, we merge these two branches to obtain the output $Z$ of of LESSM:
\begin{equation}
Z=\mathcal{C}[x'_{local}, x'_{ssm}]+x.
\end{equation}

This formulation differs from existing vision-SSM~\cite{s4nd, vmamba, vim} designs by explicitly combining global state-space modeling with local enhancement under illumination-aware guidance.

\subsection{Loss Function}
During the training process, for a given paired low-light/high-light data \(\left\{ I_{LR}^{i}, I_{HR}^{i}\right\} _{i=1}^{N}\), where \(N\) represents the number of image pairs, we use a \(L_1\) loss function to optimize the model:
\begin{equation}
\hat{\theta}= \mathop{\arg\min}\limits_{\theta} \frac{1}{N} \sum_{i=1}^{N} \Vert I_{SR}^{i} - I_{HR}^{i} \Vert_1,
\end{equation}
where \(\theta\) represents the trainable parameters; \(\Vert \cdot \Vert_1\) represents the \(L_1\) norm.

\section{Experiments}
\begin{table}[t]
\centering
\caption{
Quantitative results on LOLv1~\cite{bmvc/WeiWY018} and MIT-Adobe FiveK~\cite{vladimir2011five5k} in terms of PSNR, SSIM, and LPIPS. 
The \textbf{best} and \underline{second} results are highlighted in bold and underlined, respectively.
Higher PSNR and SSIM and lower LPIPS values mean better performance.
}
\tablestyle{13pt}{2}
\scriptsize 
\renewcommand{\arraystretch}{1.25}
\begin{tabular}{l|c|c}
\shline
\multirow{2}{*}{Method}              & LOLv1        & MIT-Adobe FiveK       
\\ \cline{2-3}
                 & PSNR/SSIM/LPIPS      & PSNR/SSIM/LPIPS        \\\shline
MF~\cite{sigpro/FuZHLDP16}             & 16.97/0.507/0.379  & 17.63/0.814/0.120      \\
SRIE~\cite{cvpr/FuZHZD16}            &11.86/0.495/0.340
  & 18.63/0.838/0.104\\
RetinexNet~\cite{bmvc/WeiWY018}            &16.77/0.425/0.473
  & 12.51/0.670/0.253\\
KinD~\cite{mm/ZhangZG19}             & 17.65/0.771/0.175
  & 16.20/0.784/0.149 \\
  
ZeroDCE~\cite{cvpr/GuoLGLHKC20} & 14.86/0.562/0.335  & 15.93/0.767/0.165
  \\
ZeroDCE++~\cite{pami/LiGL22} & 14.75/0.512/0.328  & 14.61/0.406/0.231
  \\
RUAS~\cite{cvpr/Liu0Z0L21}            & 16.40/0.503/0.270
  & 15.60/0.786/0.140
  \\
ELGAN~\cite{tip/JiangGLCFSYZW21}           & 17.48/0.651/0.322
  & 17.90/0.836/0.143
  \\
RetinexDIP~\cite{retinextip} & 11.69/0.486/0.350  & 18.67/0.813/0.116
\\
EFINet~\cite{EFINet} & 18.10/0.765/0.315  & 22.55/0.831/0.089
\\
Uformer~\cite{cvpr/WangCBZLL22}          & 18.55/0.721/0.321
  & 21.92/0.871/0.085
  \\
Restormer~\cite{cvpr/ZamirA0HK022}          & 22.37/0.816/\underline{0.141}
  & 24.92/0.911/0.058
 \\
LLFormer~\cite{aaai/WangZSLSL23}         &23.65/0.816/0.169
  & \underline{25.75}/0.923/\underline{0.045}
   \\
LLIERD~\cite{Jiang2024LightenDiffusionUL}         &20.45/0.803/0.192
  & -/-/-
   \\
SWANet~\cite{swanet} & -/-/-  & 25.21/0.896/0.055
   \\
DPRED~\cite{cviu/LiWFL24}         &-/-/-
  & 24.44/0.783/0.231
   \\
SCLM~\cite{sclm}         &19.08/0.724/0.158
  & 20.22/0.756/0.265
   \\
PPformer~\cite{cviu/DangZQ24}       &\underline{23.81}/\underline{0.821}/\textbf{0.120}
  & 24.94/0.922/-
   \\
   \shline
\textbf{MambaLIE (Ours)}           &\textbf{23.86}/\textbf{0.826}/0.153
& \textbf{25.83}/\textbf{0.928}/\textbf{0.042}
\\ \shline
\end{tabular}
\label{tab:t1}
\end{table}
\begin{table}[!t]
\centering
\caption{
Quantitative results on LOLv2-Syn~\cite{wenhua2021lolv2} and LOLv2-Real~\cite{wenhua2021lolv2}. 
}
\tablestyle{13pt}{2}
\scriptsize 
\renewcommand{\arraystretch}{1.25}
\begin{tabular}{l|c|c}
\shline
\multirow{2}{*}{Method}              & LOLv2-Syn        & LOLv2-Real       
\\ \cline{2-3}
                 & PSNR/SSIM/LPIPS      & PSNR/SSIM/LPIPS        \\\shline
MF~\cite{sigpro/FuZHLDP16}            & 17.50/0.774/0.208  & 18.72/0.509/0.240       \\
SRIE~\cite{cvpr/FuZHZD16}            & 14.50/0.664/0.248
  & 14.45/0.524/0.216\\
RetinexNet~\cite{bmvc/WeiWY018}            & 18.28/0.774/0.234
  & 16.08/0.656/0.236 \\
KinD~\cite{mm/ZhangZG19}\           & 22.01/0.904/0.273
  & 20.01/0.641/0.081  \\

KinD++~\cite{ijcv/ZhangGMLZ21}     & 21.07/0.881/0.267
  & 20.59/0.822/0.088          \\
RUAS~\cite{cvpr/Liu0Z0L21}            & 16.55/0.652/0.579  & 18.37/0.723/0.181
  \\
URetinexNet~\cite{cvpr/WuWZWYJ22}           & 24.23/0.897/\underline{0.128}
  & 21.54/0.801/\textbf{0.044}
  \\
FourLLIE~\cite{mm/WangWJ23}           & 16.85/0.852/0.423
  & 19.09/0.770/0.380
 \\
Bread~\cite{ijcv/GuoH23}          & 17.63/0.838/0.168
  & 20.83/0.822/0.095
 \\
NeRCo~\cite{iccv/YangDWLZ23}          & 19.07/0.714/0.463
  & 22.01/0.813/0.311
 \\
Retinexformer~\cite{iccv/CaiBLWTZ23}          & 24.25/\underline{0.920}/0.129
  &\underline{22.05}/0.820/0.055
 \\
CSPN~\cite{cspn}          & 21.59/0.859/0.097
  & -/-/-
  \\
PSLLIE~\cite{tcsv/LuoYYL24}          & 18.46/0.810/-
  & 17.88/0.760/-
 \\
DA-DRN~\cite{dsp/WeiLL24}          & 20.54/0.839/0.169
  & 20.73/0.794/0.313
 \\
SCLM~\cite{sclm}          & 18.81/0.630/0.164
  & 18.55/0.705/0.056
  \\
LLEMamba~\cite{corr/abs-2406-01028}                & \underline{24.27}/0.914/0.142
  & \textbf{22.20}/\textbf{0.901}/\underline{0.054}  \\   
\shline
\textbf{MambaLIE (Ours)}           & \textbf{24.67}/\textbf{0.930}/\textbf{0.055}
& \underline{22.06}/\underline{0.823}/0.187 
\\ \shline
\end{tabular}
\label{tab:t2}
\end{table}
\begin{table}[t]
\centering
\caption{Quantitative comparison on five real-world benchmark datasets in terms of no-reference image quality assessment metrics NIQE and PI. 
Lower scores in NIQE and PI  mean better quality.
}
\tablestyle{4.5pt}{1}
\scriptsize 
\renewcommand{\arraystretch}{1.25}
\begin{tabular}{l|c|c|c|c|c}
\shline
\multirow{2}{*}{Method}     &LIME&MEF&NPE&DICM&VV          
\\ \cline{2-6}
                 & NIQE/PI      & NIQE/PI     & NIQE/PI   & NIQE/PI & NIQE/PI     \\\shline
Uformer   & 4.31/3.57 &  3.85/3.25 &3.94/3.58 &3.81/ 2.97 &3.28/2.78 \\
Restormer  & 4.18/3.21 & 3.61/3.32 & 3.74/3.04 &4.23/3.26 &3.34/3.13\\
Retinexformer  &7.63/5.20  &7.67/5.30    &7.26/5.33  & 7.25/5.20  & 6.83/4.81\\
LLFormer   &6.87/4.95 &7.17/5.39&7.41/5.22&7.24/5.32 &6.23/4.55    \\
PPformer  & \textbf{3.98}/2.98 & 3.45/3.10 & 3.58/2.83 & 4.12/3.06 &3.22/2.62\\
SCLM  & 4.06/2.85 & \textbf{2.65}/2.86 & 3.57/2.61 & 3.77/2.75 &3.69/2.83\\
  \shline
\textbf{MambaLIE (Ours)}  & 4.08/\textbf{2.57} & 3.09/\textbf{2.58} & \textbf{3.54/2.60} &\textbf{3.43/2.59}& \textbf{2.32/2.45}\\
\shline
\end{tabular}
\label{tab:t3}
\vspace{-2mm}
\end{table}
In this section, we conduct extensive experiments to demonstrate the effectiveness of our proposed MambaLIE.

\subsection{Dataset and Evaluation Metrics}

\noindent\textbf{Dataset.} 
We evaluate our approach on four widely used synthetic datasets, including LOLv1~\cite{bmvc/WeiWY018}, LOLv2-real~\cite{wenhua2021lolv2}, LOLv2-syn~\cite{wenhua2021lolv2}, and MIT-AdobeFiveK~\cite{vladimir2011five5k}.
Following \cite{cvpr/GuoLGLHKC20}, we also compare the realism of enhanced performance on real datasets, including LIME~\cite{tip/GuoLL17}, MEF~\cite{kede2015MEF}, NPE~\cite{tip/WangZHL13}, DICM~\cite{tip/LeeLK13}, and VV~\cite{bmvc/WeiWY018}. 
To assess robustness under more challenging degradations, we also test our method in real-world noise and motion blur datasets.
Real-LOL-Blur~\cite{LOLBLUR}, which consists of real-world blurred low-light images taken from diverse angles and with different camera devices.
SIDD~\cite{SIDD}, which contains real-world low-light images with complex noise captured by various mobile sensors.
The training and testing sets are split in proportions of 485:15, 689:100, and 900:100 for LOL-v1, LOL-v2-real, and LOL-v2-synthetic. 
The MIT-Adobe FiveK dataset is divided into training and testing sets with 4,500 and 500 low/normal-light image pairs.
\\
\noindent \textbf{Evaluation Metrics.} 
For synthetic benchmarks, we evaluate the enhancement quality using widely used metrics, such as PSNR~\cite{PSNR_thu}, SSIM~\cite{SSIM_wang}, and LPIPS~\cite{LPIPS} with with ground-truth references.
For real-world images, where corresponding ground truth is unavailable, we use no-reference image quality assessment 
metrics, including NIQE~\cite{niqe} and PI~\cite{pi}, to evaluate the realism and perceptual quality.

\subsection{Implementation Details}
MambaLIE is trained on image patches of size 128$\times$128 with a batch size of $12$. 
Data augmentation is performed using horizontal and vertical flips. 
%
%
MambaLIE employs a 4-level encoder-decoder architecture, from stage 1 to stage 4, the number of GLEMs are $\{1, 2, 4, 8\}$.
%
%
We use the Adam optimizer~\cite{adam} with an initial learning rate of $10^{-4}$, which is gradually reduced to $10^{-6}$ using cosine annealing~\cite{loshchilov2016sgdr}. 
We train $1000$ epochs on the MIT-Adobe FiveK and $4000$ epochs on LOLv1, LOLv2-Syn, and LOLv2-Real.
All experiments are conducted on PyTorch 2.0.1 with CUDA 11.7 and an NVIDIA RTX $4090$ GPU.
\begin{figure*}[t]\footnotesize
    \centering
    \begin{minipage}[t]{0.1975\linewidth}
        \centering
        \includegraphics[width=\textwidth]{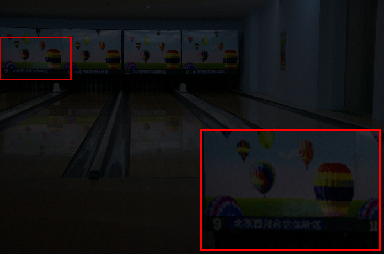}
        \subcaption*{(a) Low-light}
    \end{minipage}\hspace{-0.5mm}
    \begin{minipage}[t]{0.1975\linewidth}
        \centering
        \includegraphics[width=\textwidth]{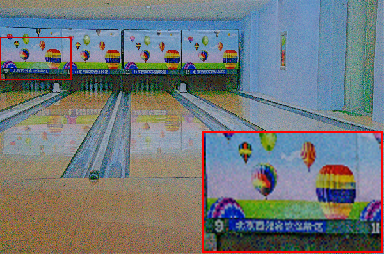}
        \subcaption*{(b) RetinexNet}
    \end{minipage}\hspace{-0.5mm}
    \begin{minipage}[t]{0.1975\linewidth}
        \centering
        \includegraphics[width=\textwidth]{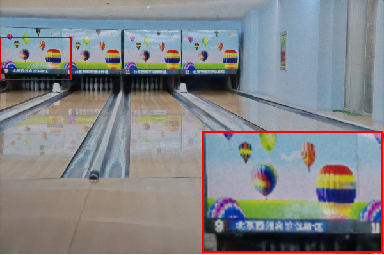}
        \subcaption*{(c) KinD}
    \end{minipage}\hspace{-0.5mm}
    \begin{minipage}[t]{0.1975\linewidth}
        \centering
        \includegraphics[width=\textwidth]{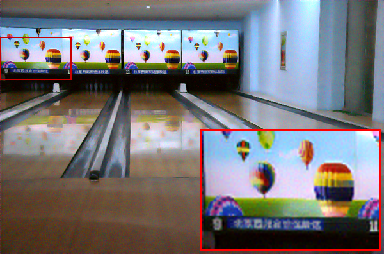}
        \subcaption*{(d) RUAS}
    \end{minipage}\hspace{-0.5mm}
    \begin{minipage}[t]{0.1975\linewidth}
        \centering
        \includegraphics[width=\textwidth]{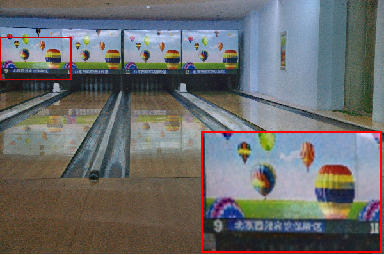}
        \subcaption*{(e) ELGAN}
    \end{minipage}


    \begin{minipage}[t]{0.1975\linewidth}
        \centering
        \includegraphics[width=\textwidth]{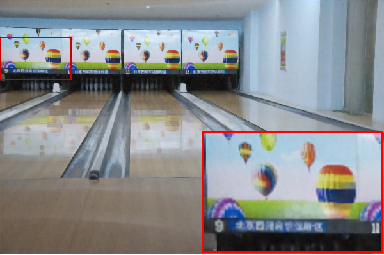}
        \subcaption*{(f) Restormer}
    \end{minipage}\hspace{-0.5mm}
    \begin{minipage}[t]{0.1975\linewidth}
        \centering
        \includegraphics[width=\textwidth]{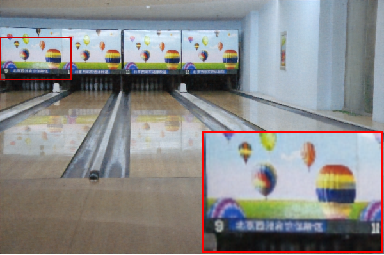}
        \subcaption*{(g) LLFormer}
    \end{minipage}\hspace{-0.5mm}
    \begin{minipage}[t]{0.1975\linewidth}
        \centering
        \includegraphics[width=\textwidth]{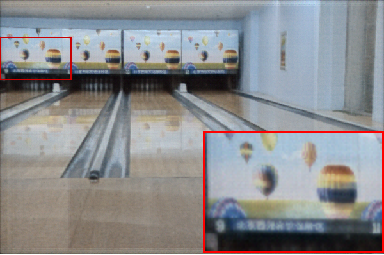}
        \subcaption*{(h) LLIEDR}
    \end{minipage}\hspace{-0.5mm}
    \begin{minipage}[t]{0.1975\linewidth}
        \centering
        \includegraphics[width=\textwidth]{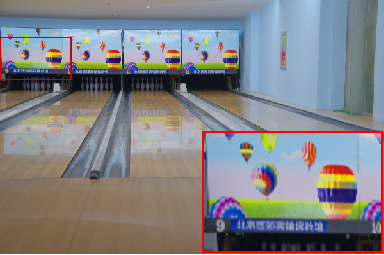}
        \subcaption*{\textbf{(i) MambaLIE (Ours)}}
    \end{minipage}\hspace{-0.5mm}
    \begin{minipage}[t]{0.1975\linewidth}
        \centering
        \includegraphics[width=\textwidth]{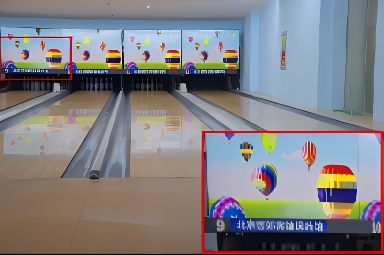}
        \subcaption*{(j) GT}
    \end{minipage}
    \caption{Visual comparison on LOLv1~\cite{bmvc/WeiWY018}. MambaLIE not only restores bright areas without overexposure but also relights dark
    areas without introducing noise and color distortion.}
    \label{fig:p3}
\end{figure*}
\begin{figure*}[t]\footnotesize
    \centering
    \begin{minipage}[t]{0.1975\linewidth}
        \centering
        \includegraphics[width=\textwidth]{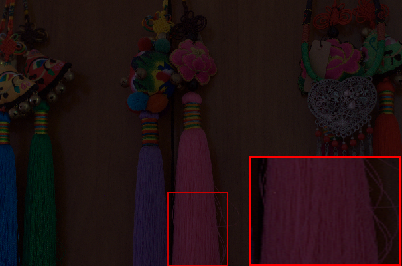}
        \subcaption*{(a) Low-light}
    \end{minipage}\hspace{-0.5mm}
    \begin{minipage}[t]{0.1975\linewidth}
        \centering
        \includegraphics[width=\textwidth]{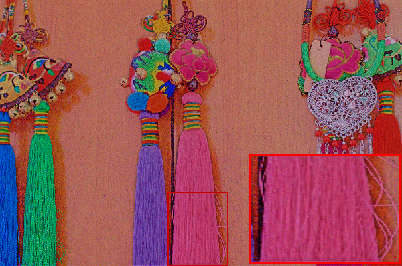}
        \subcaption*{(b) RetinexNet}
    \end{minipage}\hspace{-0.5mm}
    \begin{minipage}[t]{0.1975\linewidth}
        \centering
        \includegraphics[width=\textwidth]{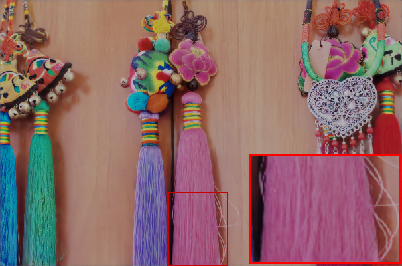}
        \subcaption*{(c) KinD}
    \end{minipage}\hspace{-0.5mm}
    \begin{minipage}[t]{0.1975\linewidth}
        \centering
        \includegraphics[width=\textwidth]{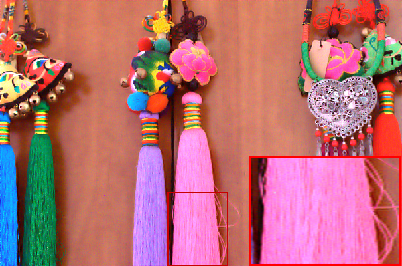}
        \subcaption*{(d) RUAS}
    \end{minipage}\hspace{-0.5mm}
    \begin{minipage}[t]{0.1975\linewidth}
        \centering
        \includegraphics[width=\textwidth]{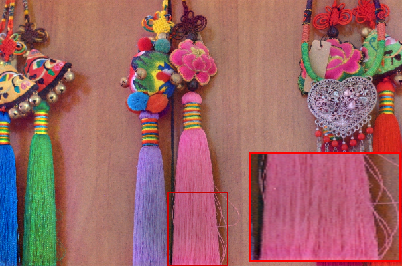}
        \subcaption*{(e) ELGAN}
    \end{minipage}


    \begin{minipage}[t]{0.1975\linewidth}
        \centering
        \includegraphics[width=\textwidth]{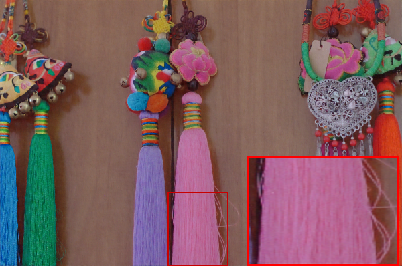}
        \subcaption*{(f) Restormer}
    \end{minipage}\hspace{-0.5mm}
    \begin{minipage}[t]{0.1975\linewidth}
        \centering
        \includegraphics[width=\textwidth]{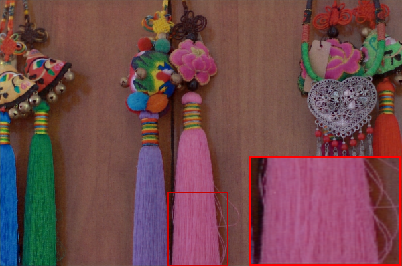}
        \subcaption*{(g) LLFormer}
    \end{minipage}\hspace{-0.5mm}
    \begin{minipage}[t]{0.1975\linewidth}
        \centering
        \includegraphics[width=\textwidth]{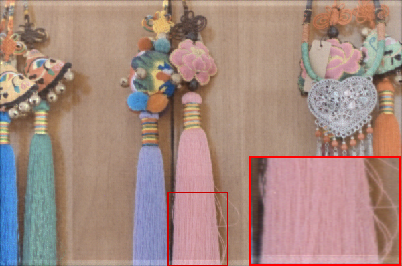}
        \subcaption*{(h) LLIEDR}
    \end{minipage}\hspace{-0.5mm}
    \begin{minipage}[t]{0.1975\linewidth}
        \centering
        \includegraphics[width=\textwidth]{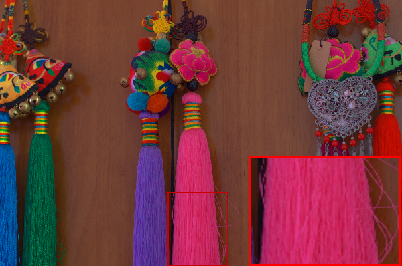}
         \subcaption*{\textbf{(i) MambaLIE (Ours)}}
    \end{minipage}\hspace{-0.5mm}
    \begin{minipage}[t]{0.1975\linewidth}
        \centering
        \includegraphics[width=\textwidth]{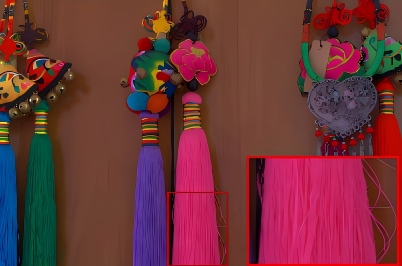}
        \subcaption*{(j) GT}
    \end{minipage}
    \caption{Results on the challenging dataset LOLv1~\cite{bmvc/WeiWY018}. Our method effectively enhances the visibility and preserves the color.}
    \label{fig:p3.01}
    \vspace{-2mm}
\end{figure*}
\begin{figure*}[t]\footnotesize
    \centering
    \begin{minipage}[t]{0.1975\linewidth}
        \centering
        \includegraphics[width=\textwidth]{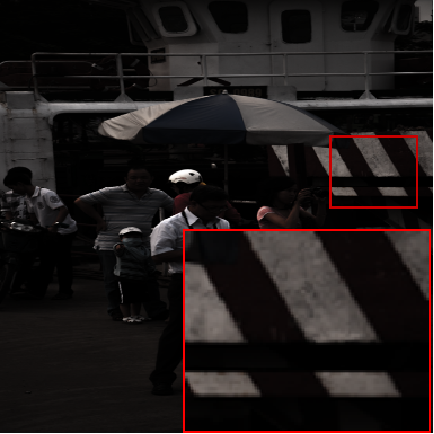}
        \subcaption*{(a) Low-light}
    \end{minipage}\hspace{-0.5mm}
    \begin{minipage}[t]{0.1975\linewidth}
        \centering
        \includegraphics[width=\textwidth]{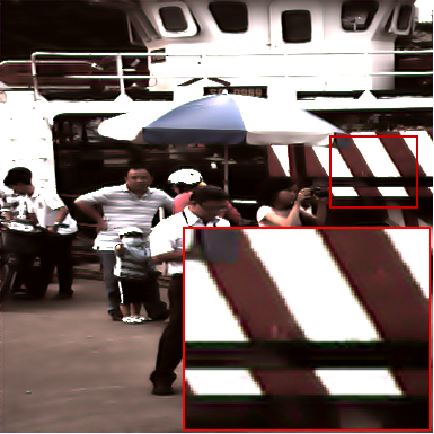}
        \subcaption*{(b) RUAS}
    \end{minipage}\hspace{-0.5mm}
    \begin{minipage}[t]{0.1975\linewidth}
        \centering
        \includegraphics[width=\textwidth]{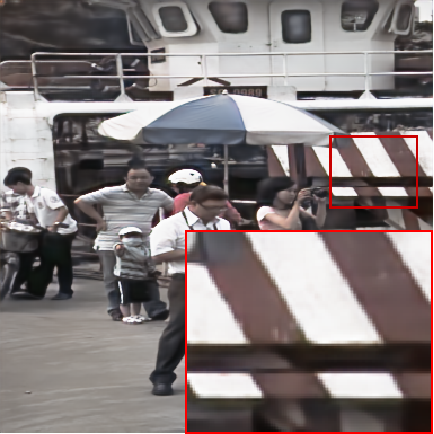}
        \subcaption*{(c) KinD}
    \end{minipage}\hspace{-0.5mm}
    \begin{minipage}[t]{0.1975\linewidth}
        \centering
        \includegraphics[width=\textwidth]{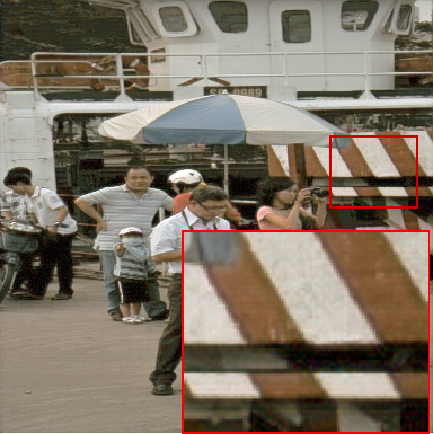}
        \subcaption*{(d) Restormer}
    \end{minipage}\hspace{-0.5mm}
    \begin{minipage}[t]{0.1975\linewidth}
        \centering
        \includegraphics[width=\textwidth]{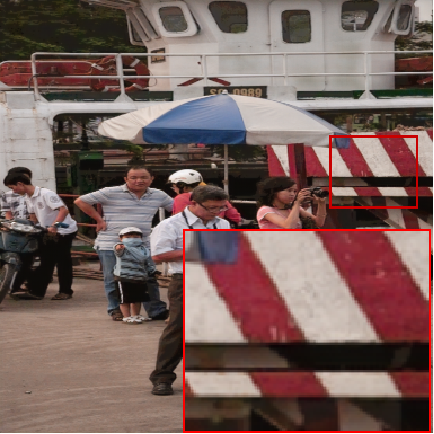}
        \subcaption*{(e) URetinexNet}
    \end{minipage}


    \begin{minipage}[t]{0.1975\linewidth}
        \centering
        \includegraphics[width=\textwidth]{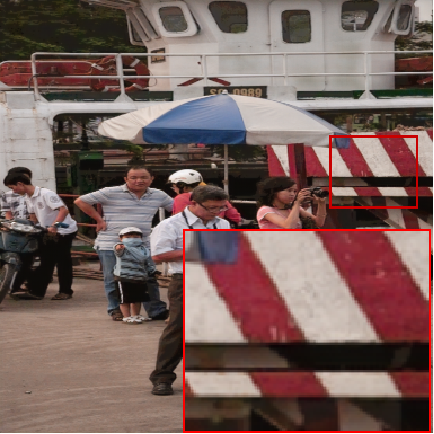}
        \subcaption*{(f) Retinexformer}
    \end{minipage}\hspace{-0.5mm}
    \begin{minipage}[t]{0.1975\linewidth}
        \centering
        \includegraphics[width=\textwidth]{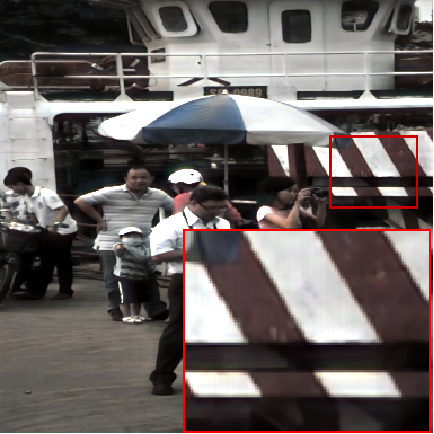}
        \subcaption*{(g) NeRCo}
    \end{minipage}\hspace{-0.5mm}
    \begin{minipage}[t]{0.1975\linewidth}
        \centering
        \includegraphics[width=\textwidth]{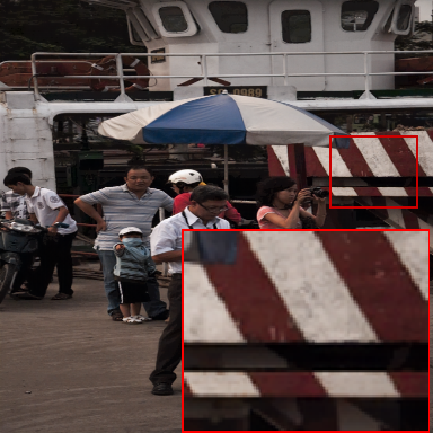}
        \subcaption*{(h) DA-DRN}
    \end{minipage}\hspace{-0.5mm}
    \begin{minipage}[t]{0.1975\linewidth}
        \centering
        \includegraphics[width=\textwidth]{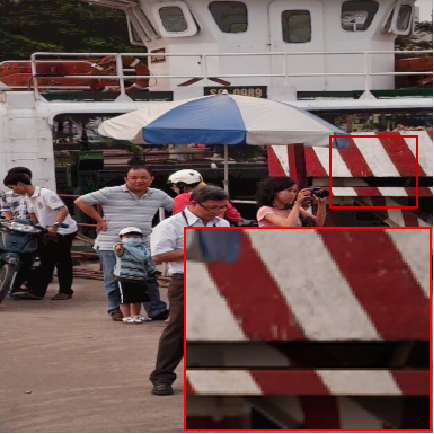}
         \subcaption*{\textbf{(i) MambaLIE (Ours)}}
    \end{minipage}\hspace{-0.5mm}
    \begin{minipage}[t]{0.1975\linewidth}
        \centering
        \includegraphics[width=\textwidth]{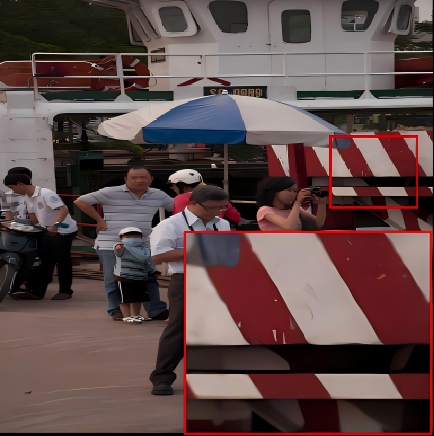}
        \subcaption*{(j) GT}
    \end{minipage}
    \caption{Results on the challenging dataset LOLv2-syn~\cite{wenhua2021lolv2}.
    Our method is capable of improving both brightness and detail, resulting in a more vivid and clearer image with enhanced visual quality.}
    \label{fig:p3.1}
\end{figure*}

\begin{figure*}[t]\footnotesize
    \centering
    \begin{minipage}[t]{0.1975\linewidth}
        \centering
        \includegraphics[width=\textwidth]{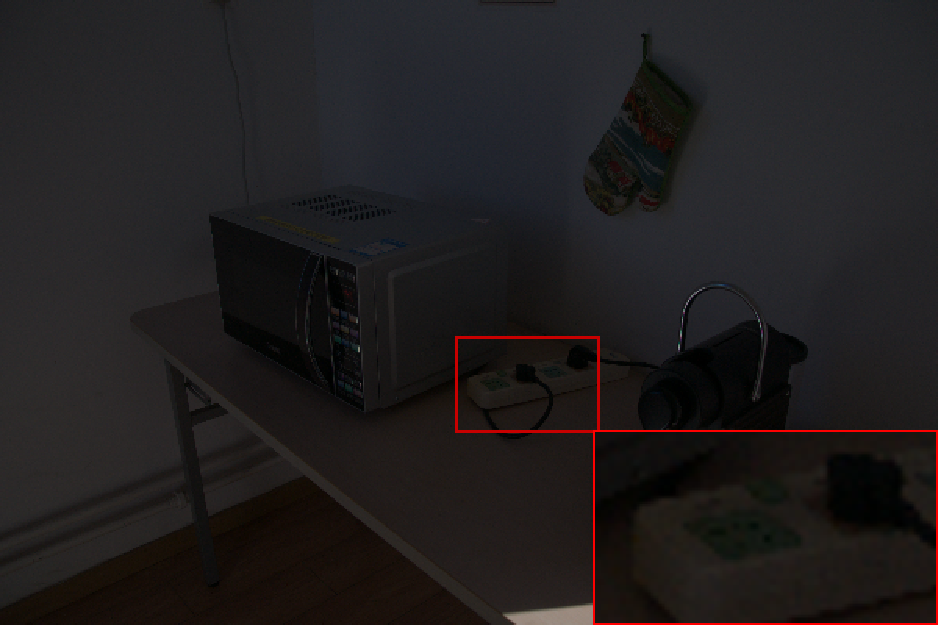}
        \subcaption*{(a) Low-light}
    \end{minipage}\hspace{-0.5mm}
    \begin{minipage}[t]{0.1975\linewidth}
        \centering
        \includegraphics[width=\textwidth]{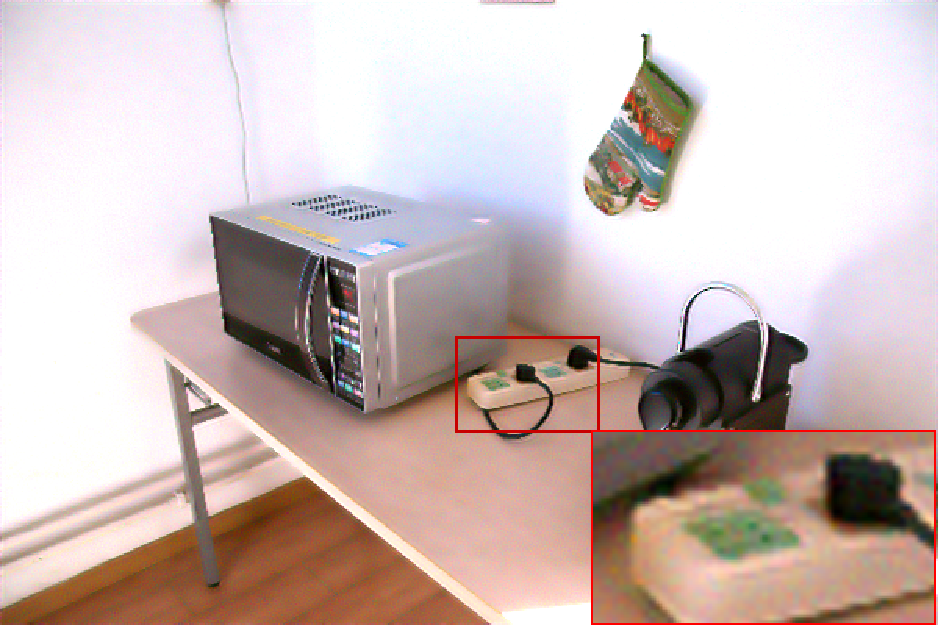}
        \subcaption*{(b) RUAS}
    \end{minipage}\hspace{-0.5mm}
    \begin{minipage}[t]{0.1975\linewidth}
        \centering
        \includegraphics[width=\textwidth]{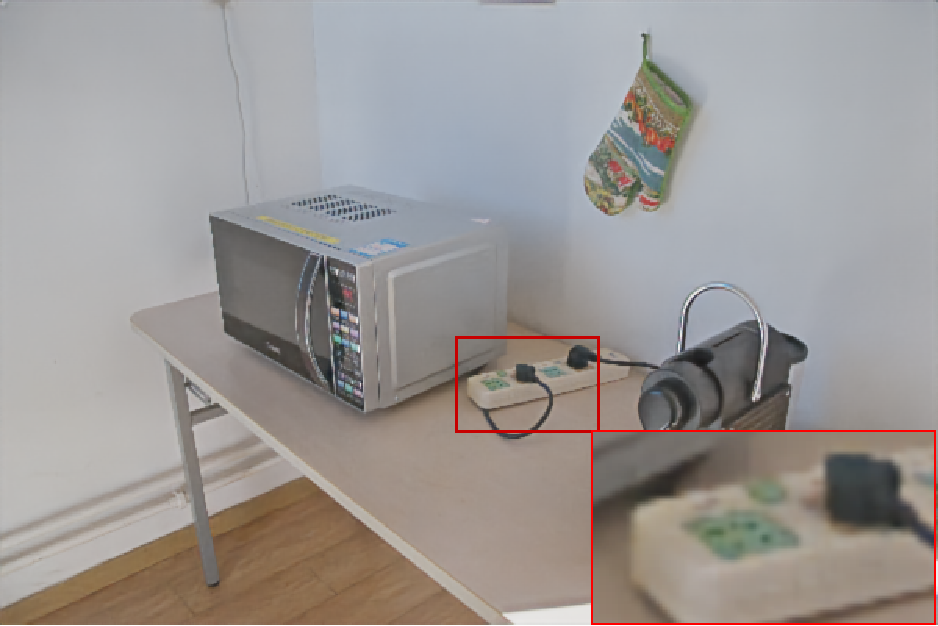}
        \subcaption*{(c) KinD}
    \end{minipage}\hspace{-0.5mm}
    \begin{minipage}[t]{0.1975\linewidth}
        \centering
        \includegraphics[width=\textwidth]{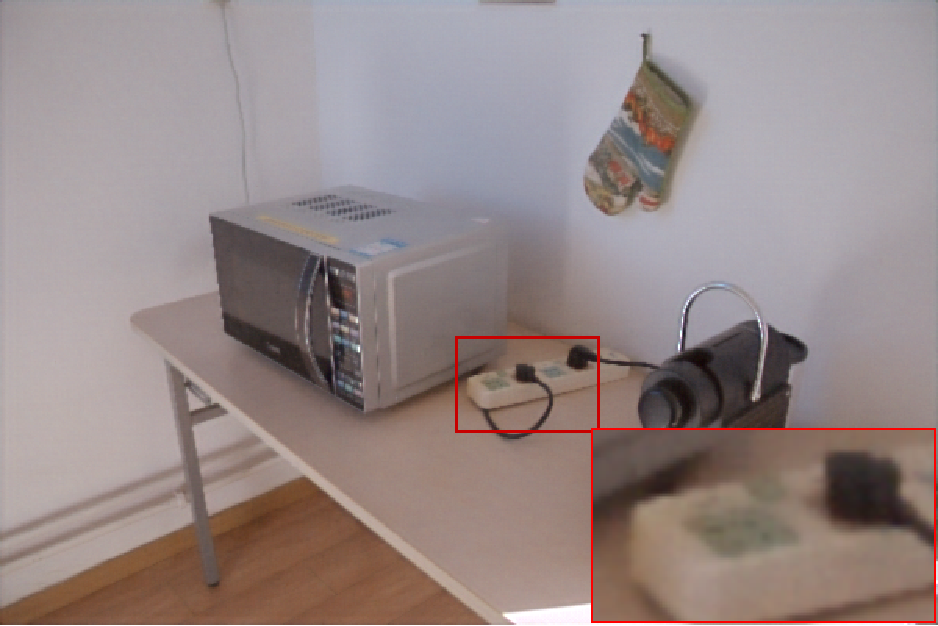}
        \subcaption*{(d) Restormer}
    \end{minipage}\hspace{-0.5mm}
    \begin{minipage}[t]{0.1975\linewidth}
        \centering
        \includegraphics[width=\textwidth]{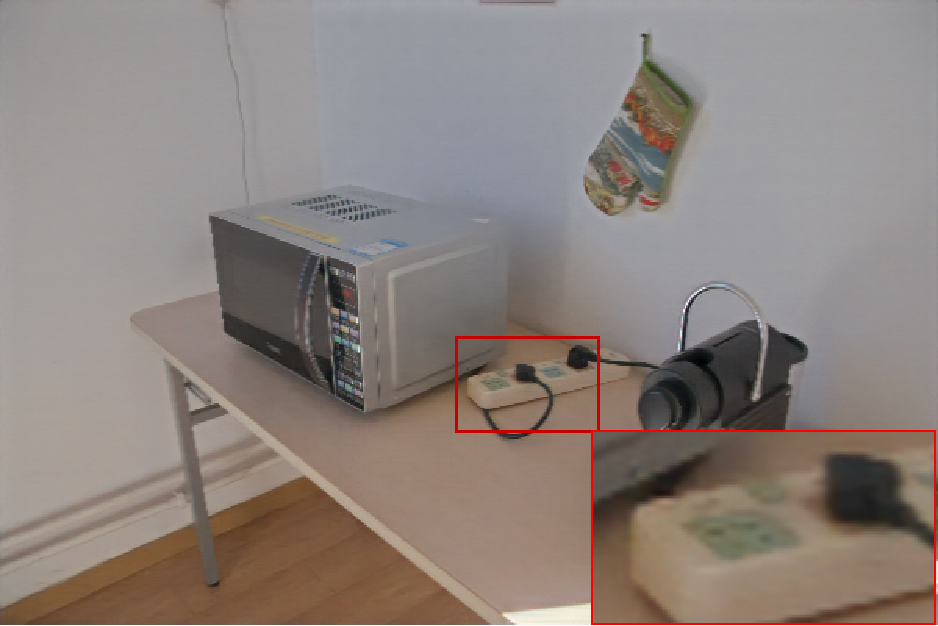}
        \subcaption*{(e) URetinexNet}
    \end{minipage}


    \begin{minipage}[t]{0.1975\linewidth}
        \centering
        \includegraphics[width=\textwidth]{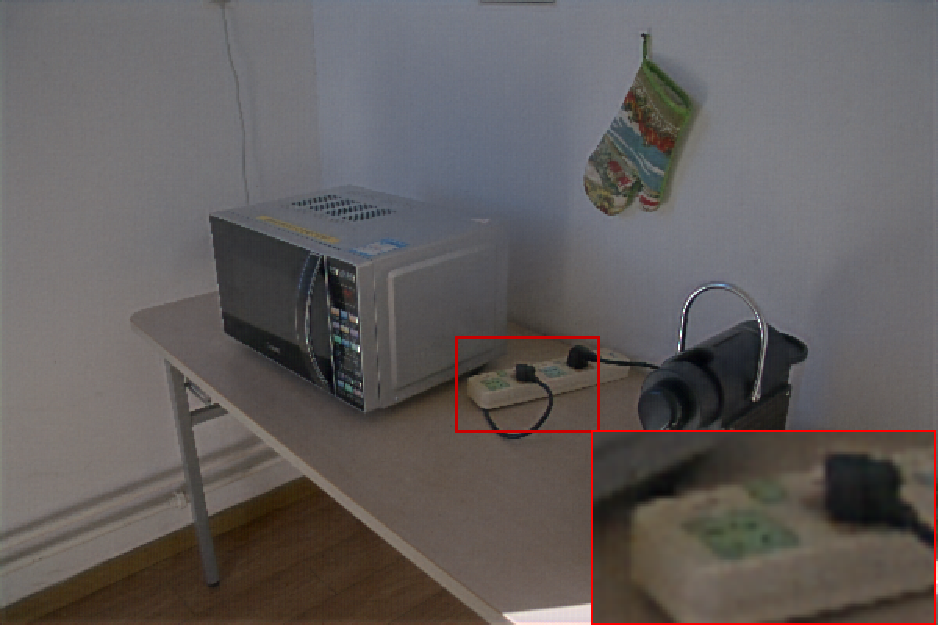}
        \subcaption*{(f) Retinexformer}
    \end{minipage}\hspace{-0.5mm}
    \begin{minipage}[t]{0.1975\linewidth}
        \centering
        \includegraphics[width=\textwidth]{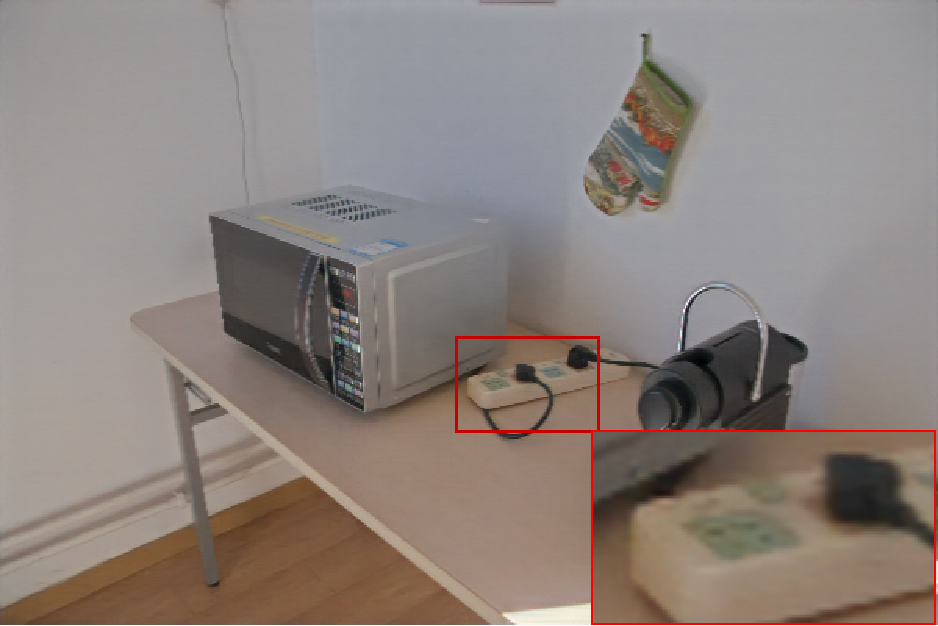}
        \subcaption*{(g) NeRCo}
    \end{minipage}\hspace{-0.5mm}
    \begin{minipage}[t]{0.1975\linewidth}
        \centering
        \includegraphics[width=\textwidth]{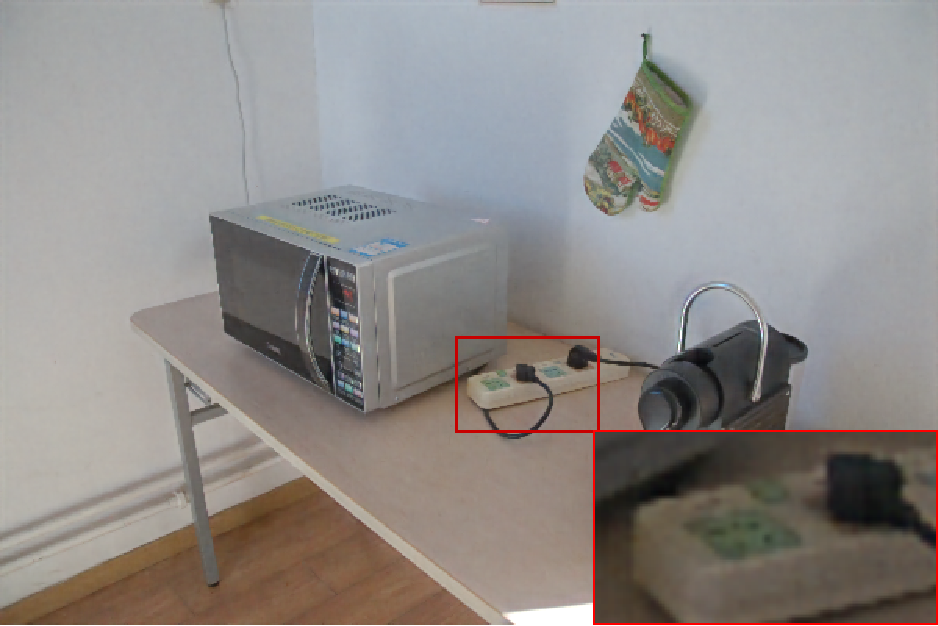}
        \subcaption*{(h) DA-DRN}
    \end{minipage}\hspace{-0.5mm}
    \begin{minipage}[t]{0.1975\linewidth}
        \centering
        \includegraphics[width=\textwidth]{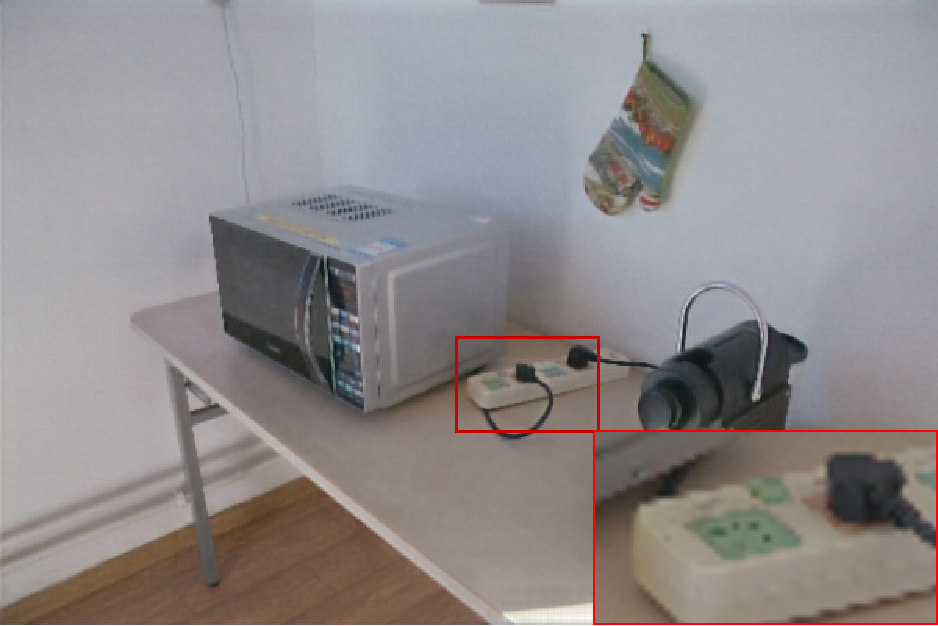}
        \subcaption*{\textbf{(i) MambaLIE (Ours)}}
    \end{minipage}\hspace{-0.5mm}
    \begin{minipage}[t]{0.1975\linewidth}
        \centering
        \includegraphics[width=\textwidth]{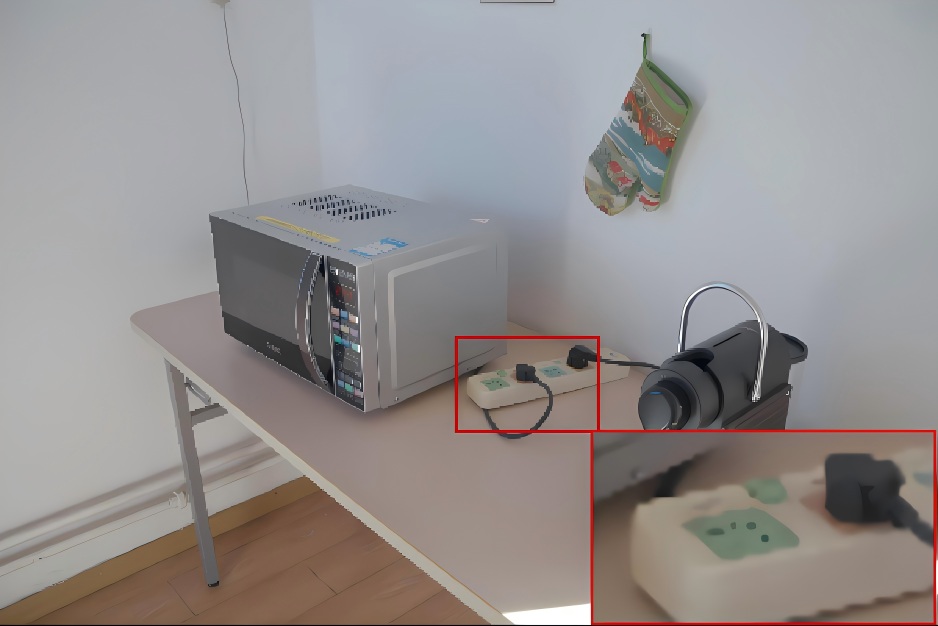}
        \subcaption*{(j) GT}
    \end{minipage}
    \caption{Visual comparison on LOLv2-Real~\cite{wenhua2021lolv2}. MambaLIE effectively preserves the brightness contrast between objects, ensuring that the distinction in lighting remains clear and natural throughout the image.}
    \label{fig:p4}
\end{figure*}
\begin{figure*}[t]\footnotesize
    \centering
    \begin{minipage}[t]{0.1975\linewidth}
        \centering
        \includegraphics[width=\textwidth]{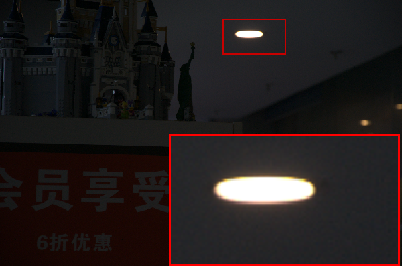}
        \subcaption*{(a) Low-light}
    \end{minipage}\hspace{-0.5mm}
    \begin{minipage}[t]{0.1975\linewidth}
        \centering
        \includegraphics[width=\textwidth]{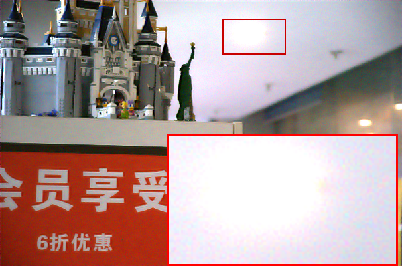}
        \subcaption*{(b) RUAS}
    \end{minipage}\hspace{-0.5mm}
    \begin{minipage}[t]{0.1975\linewidth}
        \centering
        \includegraphics[width=\textwidth]{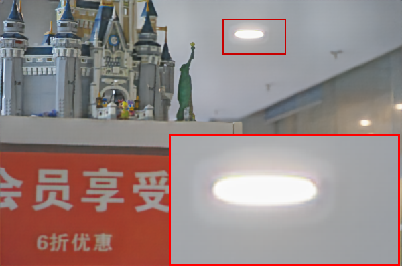}
        \subcaption*{(c) KinD}
    \end{minipage}\hspace{-0.5mm}
    \begin{minipage}[t]{0.1975\linewidth}
        \centering
        \includegraphics[width=\textwidth]{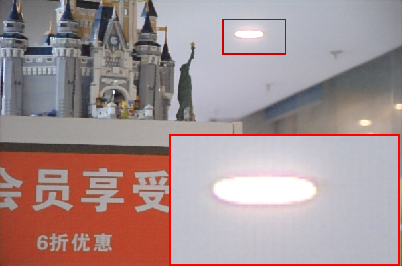}
        \subcaption*{(d) Restormer}
    \end{minipage}\hspace{-0.5mm}
    \begin{minipage}[t]{0.1975\linewidth}
        \centering
        \includegraphics[width=\textwidth]{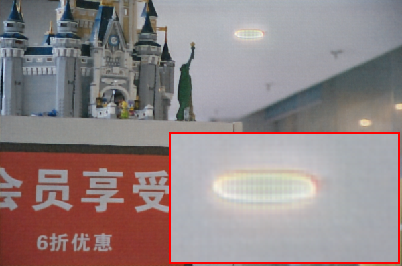}
        \subcaption*{(e) URetinexNet}
    \end{minipage}


    \begin{minipage}[t]{0.1975\linewidth}
        \centering
        \includegraphics[width=\textwidth]{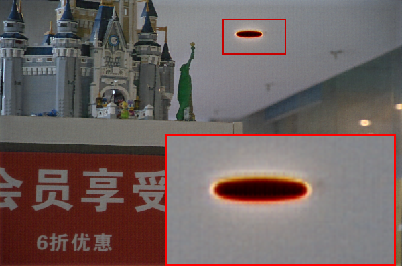}
        \subcaption*{(f) Retinexformer}
    \end{minipage}\hspace{-0.5mm}
    \begin{minipage}[t]{0.1975\linewidth}
        \centering
        \includegraphics[width=\textwidth]{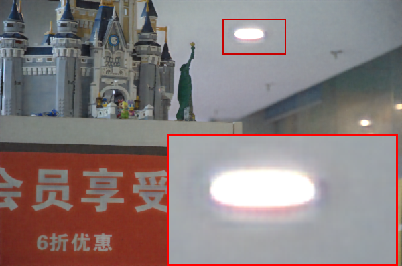}
        \subcaption*{(g) NeRCo}
    \end{minipage}\hspace{-0.5mm}
    \begin{minipage}[t]{0.1975\linewidth}
        \centering
        \includegraphics[width=\textwidth]{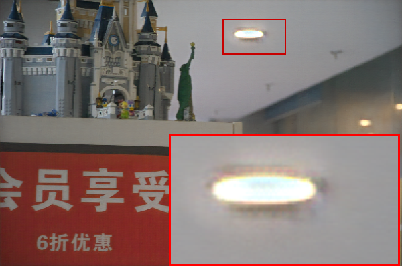}
        \subcaption*{(h) DA-DRN}
    \end{minipage}\hspace{-0.5mm}
    \begin{minipage}[t]{0.1975\linewidth}
        \centering
        \includegraphics[width=\textwidth]{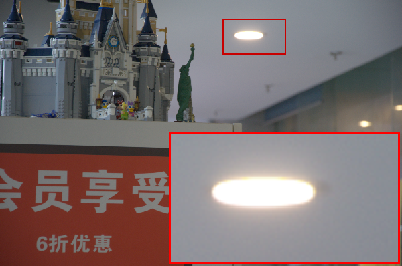}
        \subcaption*{\textbf{(i) MambaLIE (Ours)}}
    \end{minipage}\hspace{-0.5mm}
    \begin{minipage}[t]{0.1975\linewidth}
        \centering
        \includegraphics[width=\textwidth]{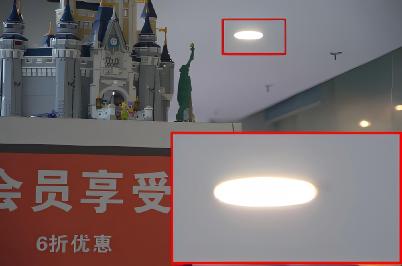}
        \subcaption*{(j) GT}
    \end{minipage}
    \caption{Results on the challenging dataset LOLv2-real~\cite{wenhua2021lolv2}.
    Our MambaLIE method enhances the image effectively while ensuring that overexposure is avoided.}
    \label{fig:p41}
\end{figure*}

\subsection{Quantitative Results}
We evaluate the performance of our MambaLIE against SOTA LIE methods, as shown in Table~\ref{tab:t1} and Table~\ref{tab:t2}.
Our MambaLIE achieves competitive performance according to PSNR, SSIM, and LPIPS.
Higher PSNR values demonstrate that our MambaLIE effectively suppresses artifacts and recovers color information. 
Better SSIM values show that our MambaLIE better preserves structural information with richer details. 
Furthermore, MambaLIE achieves the best performance in LPIPS, which
is designed to align with human perception, indicating that our method produces results that are more perceptually pleasing.
%

%
We further compare the realism quality of enhanced results in Table~\ref{tab:t3} in terms of NIQE and PI on the LIME~\cite{tip/GuoLL17}, MEF~\cite{kede2015MEF}, NPE~\cite{tip/WangZHL13}, DICM~\cite{tip/LeeLK13}, and VV~\cite{bmvc/WeiWY018} datasets. 
We test these five datasets using a model trained on LOLv1.
One can observe that our MambaLIE is able to achieve excellent performance compared to SOTA approaches, such as the Transformer-based method PPformer~\cite{cviu/DangZQ24}.
The results reveal that our MambaLIE is able to restore results with superior perceptual quality and realism.
\subsection{Qualitative Results on Synthetic Benchmarks.}
Fig.~\ref{fig:p3}, Fig.~\ref{fig:p3.01} and Fig.~\ref{fig:p3.1} present a comprehensive qualitative comparison of our method against other techniques on synthetic benchmark datasets. These figures collectively demonstrate the effectiveness of MambaLIE in tackling various image enhancement challenges.
Fig.~\ref{fig:p3} highlights MambaLIE's ability to recover more accurate and precise color information compared to competing methods. This accurate color restoration is critical in ensuring that the enhanced images maintain their natural look and feel, even under challenging synthetic conditions.
Fig.~\ref{fig:p3.01} shows that our method is shown to significantly improve visibility while simultaneously preserving the original color fidelity. This balance between enhancing clarity and maintaining true-to-life colors sets MambaLIE apart from other approaches that often introduce artifacts or color distortions during the enhancement process.
Fig.~\ref{fig:p3.1} showcases MambaLIE's strength in producing not only superior brightness but also in restoring fine details. These improvements are particularly evident in darker regions of the images, where competing methods often struggle to illuminate without losing texture or introducing noise. MambaLIE's ability to strike this balance highlights its robustness in handling both brightness and detail restoration, ensuring a more comprehensive image enhancement.
\subsection{Qualitative Results on Real-World Benchmarks.}
\begin{figure*}[!t]\footnotesize
\begin{center}
\begin{tabular}{cccccc}
\hspace{-1.5mm}\includegraphics[width = 0.163\linewidth]{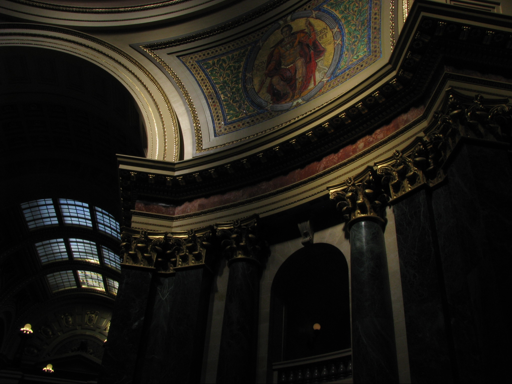} &\hspace{-4.5mm}
\includegraphics[width = 0.163\linewidth]{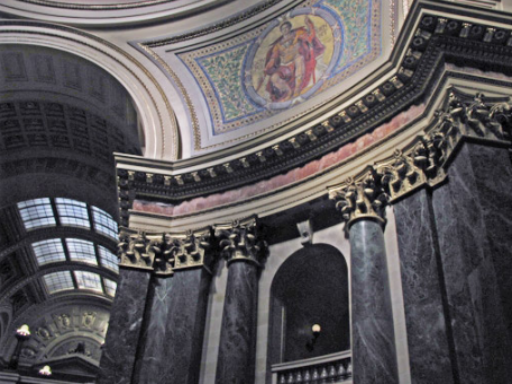}  &\hspace{-4.5mm}
\includegraphics[width = 0.163\linewidth]{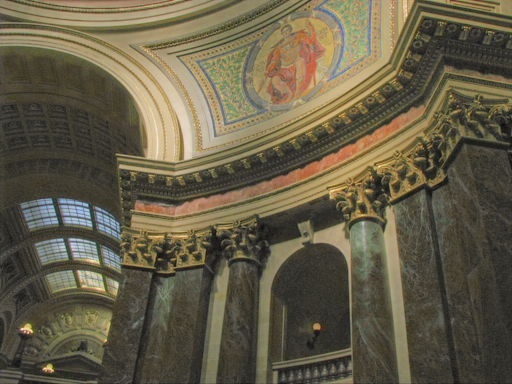} &\hspace{-4.5mm}
\includegraphics[width = 0.163\linewidth]{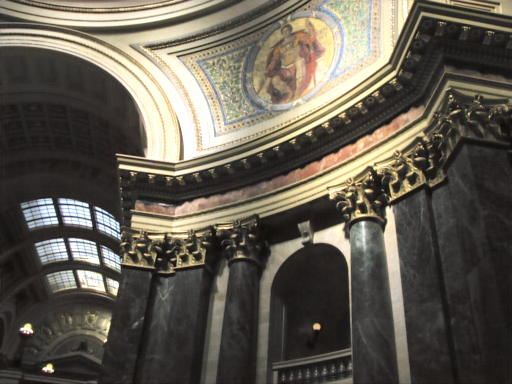} &\hspace{-4.5mm}

\includegraphics[width = 0.163\linewidth]{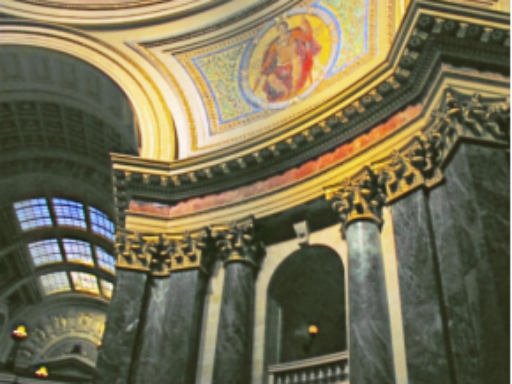} &\hspace{-4.5mm}
\includegraphics[width = 0.163\linewidth]{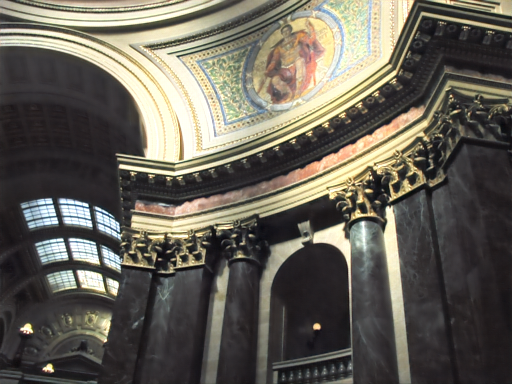} 
\\
\hspace{-1.5mm}\includegraphics[width = 0.163\linewidth]{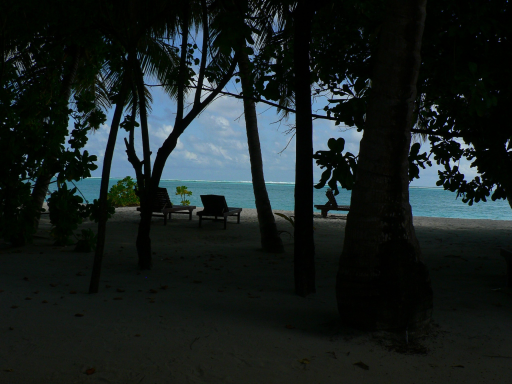} &\hspace{-4.5mm}
\includegraphics[width = 0.163\linewidth]{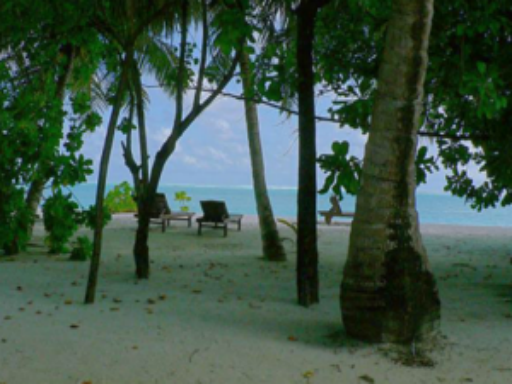} &\hspace{-4.5mm}
\includegraphics[width = 0.163\linewidth]{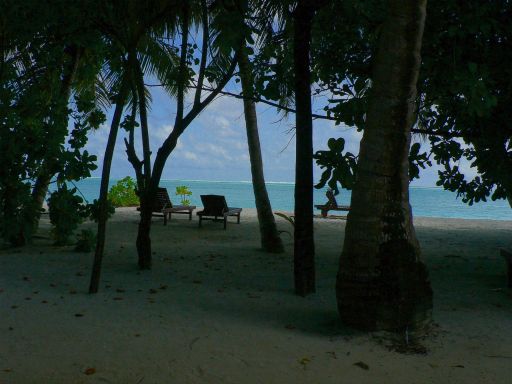} &\hspace{-4.5mm}
\includegraphics[width = 0.163\linewidth]{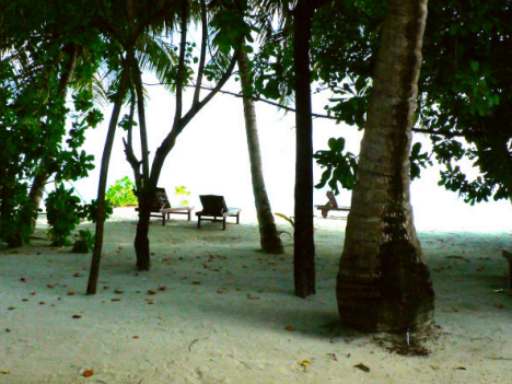} &\hspace{-4.5mm}
\includegraphics[width = 0.163\linewidth]{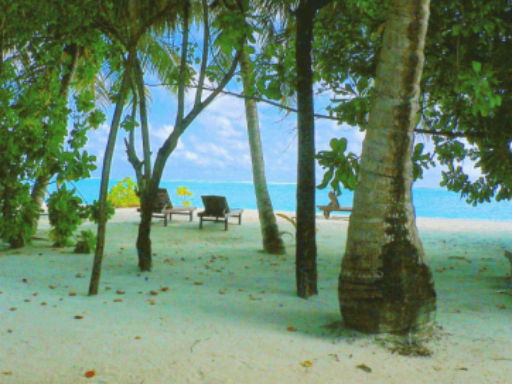}  &\hspace{-4.5mm}
\includegraphics[width = 0.163\linewidth]{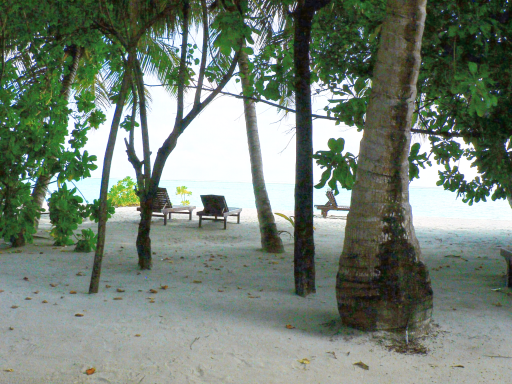} 
\\
\hspace{-1.5mm}(a) Low-light   &\hspace{-4.5mm} (b) ZeroDCE++ &\hspace{-4.5mm} (c) Retinexformer&\hspace{-4.5mm} (d) LLFormer&\hspace{-4.5mm} (e)  PSLLIE&\hspace{-4.5mm} (f) \textbf{Ours}
\end{tabular}
\end{center}
\vspace{-2mm}
\caption{Visual comparison of real-world images from MEF~\cite{kede2015MEF} and VV~\cite{bmvc/WeiWY018}. Our MambaLIE is able to produce a better visual effect with more natural colors.
}
\label{fig:p5}
\end{figure*}

\begin{figure*}[!h]\footnotesize
\begin{center}
\begin{tabular}{ccccc}
\includegraphics[width = 0.1975\linewidth]{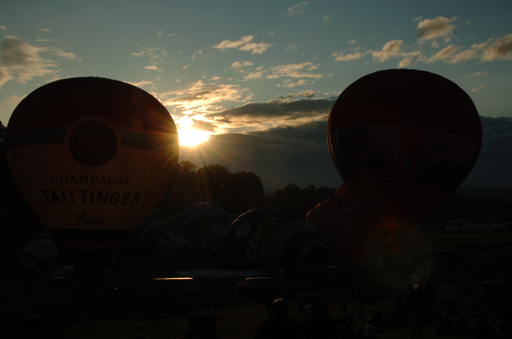} &\hspace{-4.5mm}
\includegraphics[width = 0.1975\linewidth]{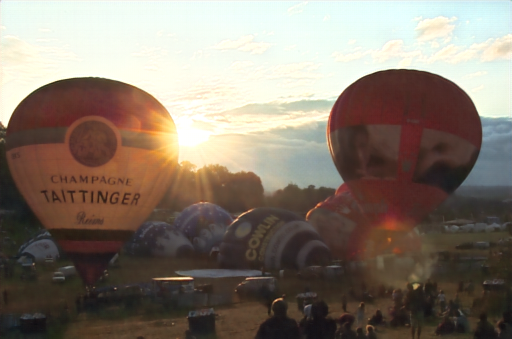} &\hspace{-4.5mm}
\includegraphics[width = 0.1975\linewidth]{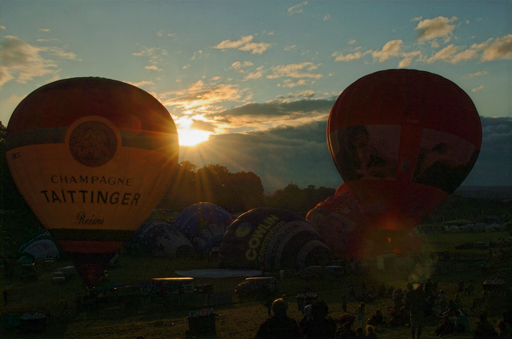} &\hspace{-4.5mm}
\includegraphics[width = 0.1975\linewidth]{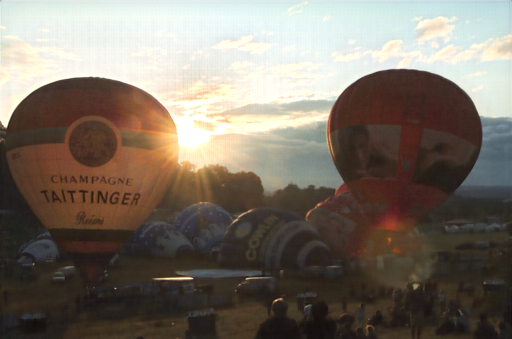} &\hspace{-4.5mm}
\includegraphics[width = 0.1975\linewidth]{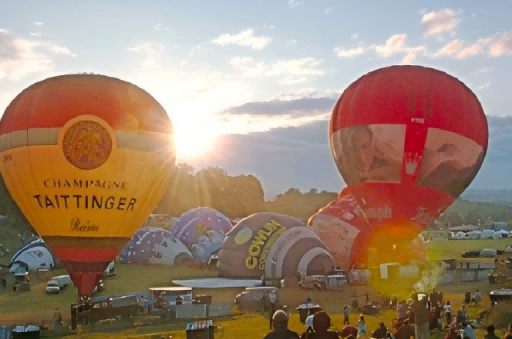}  
\\
\includegraphics[width = 0.1975\linewidth]{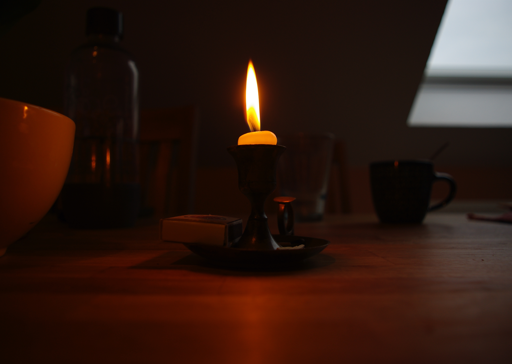} &\hspace{-4.5mm}
\includegraphics[width = 0.1975\linewidth]{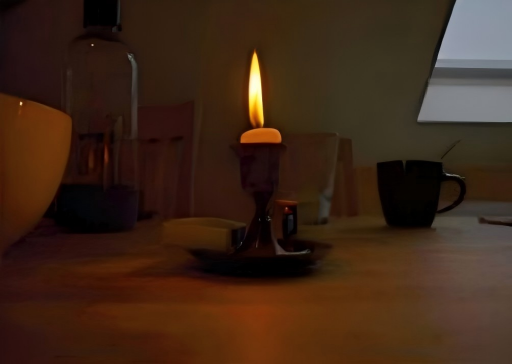}  &\hspace{-4.5mm}
\includegraphics[width = 0.1975\linewidth]{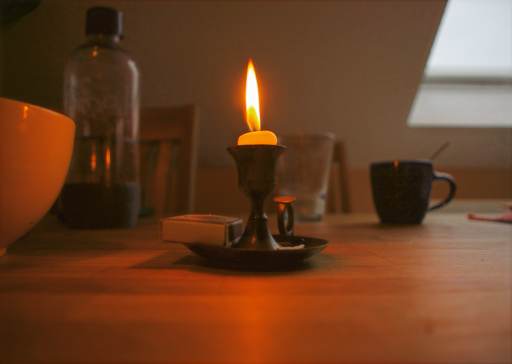} &\hspace{-4.5mm}
\includegraphics[width = 0.1975\linewidth]{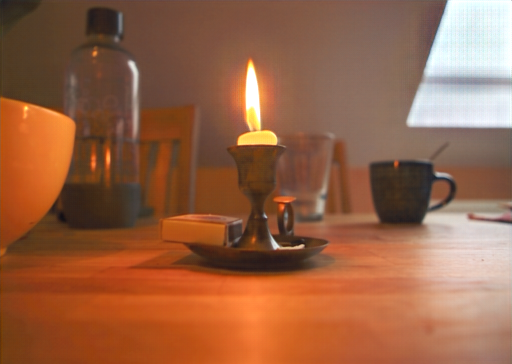} &\hspace{-4.5mm}

\includegraphics[width = 0.1975\linewidth]{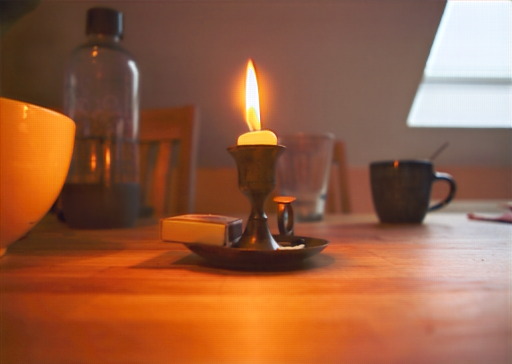} 
\\
(a) Low-Light   &\hspace{-4.5mm} (b) ZeroDCE++ &\hspace{-4.5mm} (c) Retinexformer&\hspace{-4.5mm} (d) LLFormer&\hspace{-4.5mm} (e)  \textbf{Ours}
\end{tabular}
\end{center}
\vspace{-2mm}
\caption{Visual comparison of real-world images from MEF~\cite{kede2015MEF} and LIME~\cite{tip/LeeLK13}. Our method not only achieves a more visually pleasing result but also enhances the color naturalness, offering a superior visual effect compared to other approaches.
}
\label{fig:p62}
\end{figure*}

Fig.~\ref{fig:p4} showcases the impressive capability of MambaLIE to enhance images by maintaining consistent color fidelity and minimizing visual artifacts. This consistency is crucial in delivering high-quality image enhancement, particularly in challenging lighting conditions.
Fig.~\ref{fig:p41} further illustrates the strength of our method, effectively enhancing the image without the common issue of overexposure, thus preserving the natural balance of light and shadow.
Fig.~\ref{fig:p5}, a comparative analysis on real-world benchmark datasets is presented, highlighting the performance of our method against state-of-the-art (SOTA) algorithms. The comparison reveals the superior visual quality achieved by MambaLIE, especially in terms of color reproduction and artifact suppression.
One the other hand, Fig.~\ref{fig:p62} demonstrates that our approach not only produces visually appealing results but also excels in enhancing color naturalness under highly non-uniform lighting conditions.
This ensures a more realistic visual effect when compared to other methods, offering both aesthetic and technical improvements.
%
%
%
Additionally, while existing methods often struggle with challenges such as insufficient illumination and noticeable color distortions, our MambaLIE method addresses these issues head-on. 
It not only improves brightness but also accurately preserves the original color integrity, ensuring that fine details are restored, leading to a more comprehensive image enhancement solution.

\begin{figure*}[!t]\footnotesize
\begin{center}
\begin{tabular}{ccccc}
\hspace{-1.5mm}\includegraphics[width = 0.1975\linewidth]{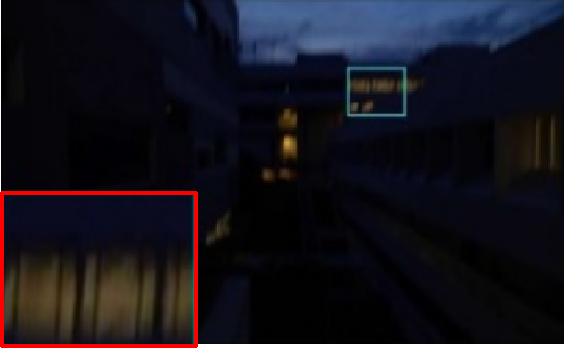} &\hspace{-4.5mm}
\includegraphics[width = 0.1975\linewidth]{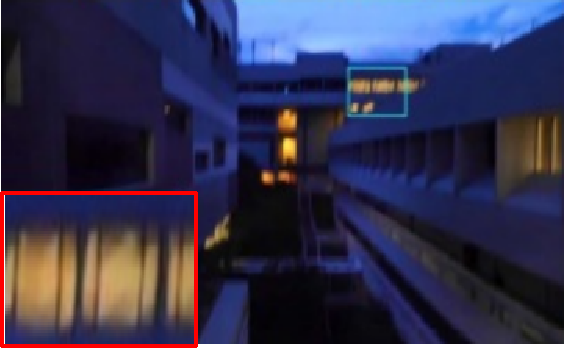}  &\hspace{-4.5mm}
\includegraphics[width = 0.1975\linewidth]{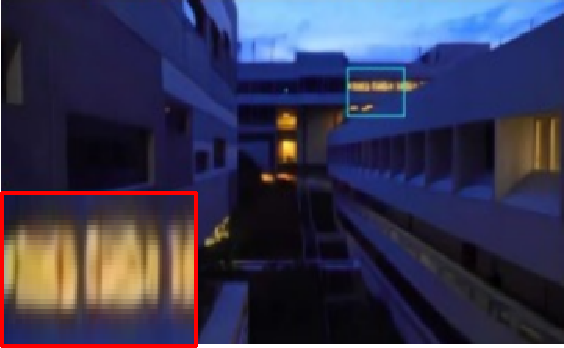} &\hspace{-4.5mm}
\includegraphics[width = 0.1975\linewidth]{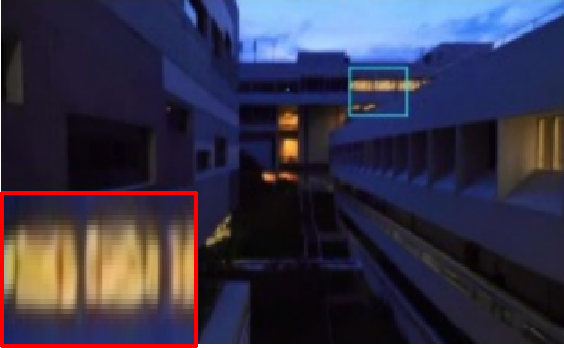} &\hspace{-4.5mm}

\includegraphics[width = 0.1975\linewidth]{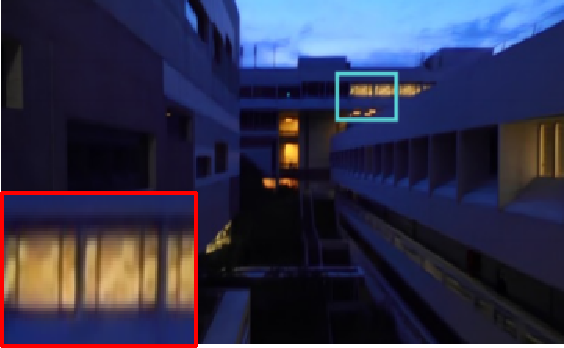} 
\\
\hspace{-1.5mm}\includegraphics[width = 0.1975\linewidth]{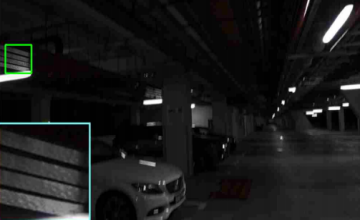} &\hspace{-4.5mm}
\includegraphics[width = 0.1975\linewidth]{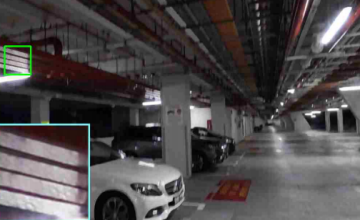} &\hspace{-4.5mm}
\includegraphics[width = 0.1975\linewidth]{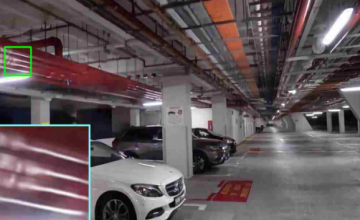} &\hspace{-4.5mm}
\includegraphics[width = 0.1975\linewidth]{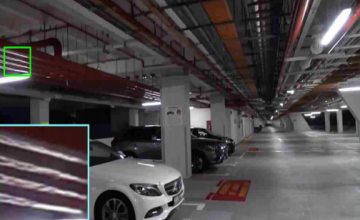} &\hspace{-4.5mm}
\includegraphics[width = 0.1975\linewidth]{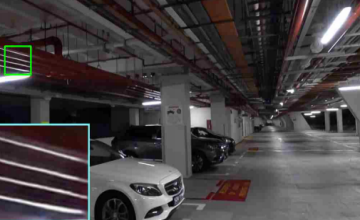}  
\\
\hspace{-1.5mm}(a) Low-light   &\hspace{-4.5mm} (b) Restormer&\hspace{-4.5mm} (c) Retinexformer&\hspace{-4.5mm} (d) LLFormer&\hspace{-4.5mm} (e)  \textbf{Ours}
\end{tabular}
\end{center}
\vspace{-2mm}
\caption{Visual results on blurred low-light images from the Real-LOL-Blur~\cite{LOLBLUR}.
MambaLIE generates much sharper images with visually pleasing results, indicating robustness to real-world blur in low-light conditions.}
\label{fig:xin1}
\end{figure*}

\begin{figure*}[!t]\footnotesize
\begin{center}
\begin{tabular}{ccccc}
\hspace{-1.5mm}\includegraphics[width = 0.1975\linewidth]{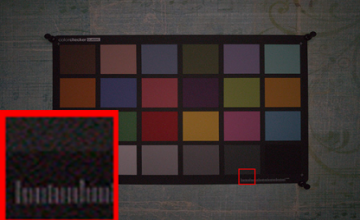} &\hspace{-4.5mm}
\includegraphics[width = 0.1975\linewidth]{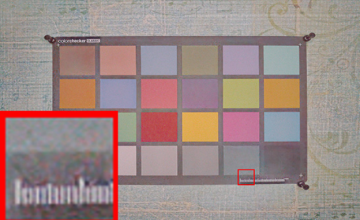}  &\hspace{-4.5mm}
\includegraphics[width = 0.1975\linewidth]{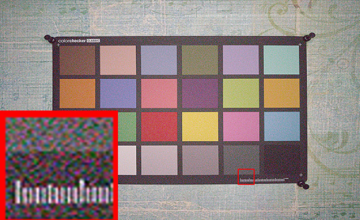} &\hspace{-4.5mm}
\includegraphics[width = 0.1975\linewidth]{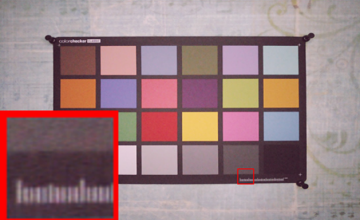} &\hspace{-4.5mm}

\includegraphics[width = 0.1975\linewidth]{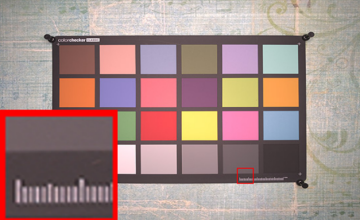} 
\\
\hspace{-1.5mm}\includegraphics[width = 0.1975\linewidth]{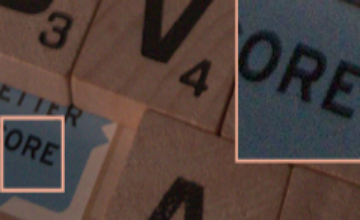} &\hspace{-4.5mm}
\includegraphics[width = 0.1975\linewidth]{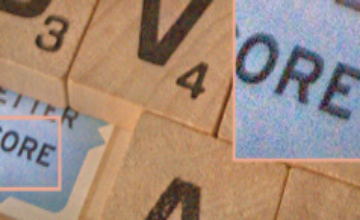} &\hspace{-4.5mm}
\includegraphics[width = 0.1975\linewidth]{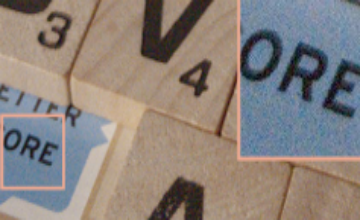} &\hspace{-4.5mm}
\includegraphics[width = 0.1975\linewidth]{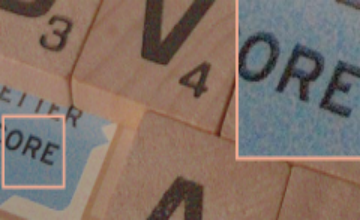} &\hspace{-4.5mm}
\includegraphics[width = 0.1975\linewidth]{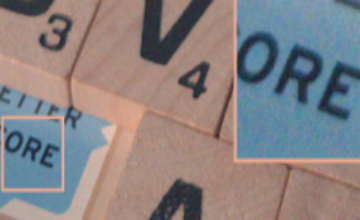}  
\\
\hspace{-1.5mm}(a) Low-light   &\hspace{-4.5mm} (b) Restormer&\hspace{-4.5mm} (c) Retinexformer&\hspace{-4.5mm} (d) LLFormer&\hspace{-4.5mm} (e)  \textbf{Ours}
\end{tabular}
\end{center}
\vspace{-2mm}
\caption{Visual results on noisy low-light scenes from the SIDD~\cite{SIDD}. 
MambaLIE produces clean images
while preserving fine textures, indicating robustness to real-world noise in low-light conditions.}
\label{fig:xin2}
\end{figure*}

\begin{table}[!t]
\centering
\tablestyle{1pt}{1}
\scriptsize 
\caption{Quantitative comparison in terms of NIQE/MUSIQ on two real-world datasets. Our method achieves the best NIQE and MUSIQ.}
\renewcommand{\arraystretch}{1.5}
\begin{tabular}{c|cccc}
\hline
\multirow{2}{*}{Datasets} & Resotmer~\cite{cvpr/ZamirA0HK022}  & Retinexformer~\cite{iccv/CaiBLWTZ23} & LLFormer~\cite{aaai/WangZSLSL23} & Ours \\
\cline{2-5}
 & NIQE$\downarrow$/MUSIQ$\uparrow$ & NIQE$\downarrow$/MUSIQ$\uparrow$ & NIQE$\downarrow$/MUSIQ$\uparrow$ & NIQE$\downarrow$/MUSIQ$\uparrow$  \\
\hline
Real-LOL-Blur   &  5.02/36.62  & 4.86/37.22 & 4.65/38.56& 4.23/40.21  \\ 
SIDD   &  4.23/34.25  & 4.15/36.46 & 4.02/38.55& 3.98/39.63  \\
\hline
\end{tabular}
\label{tab:xin1}
\vspace{-2mm}
\end{table}

\begin{figure}[t]\footnotesize
\begin{center}
\begin{tabular}{ccc}  
\hspace{-1.5mm}\includegraphics[width = 0.325\linewidth]{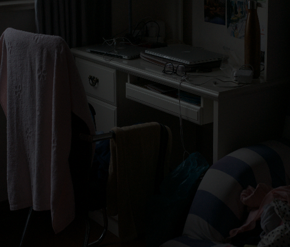} &\hspace{-4.5mm}
\includegraphics[width = 0.325\linewidth]{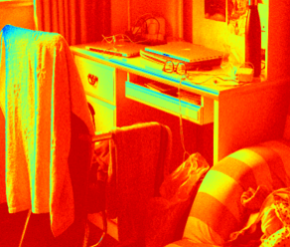} &\hspace{-4.5mm}
\includegraphics[width = 0.325\linewidth]{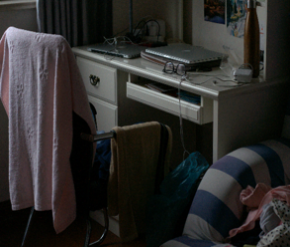} 
\\
\hspace{-1.5mm}(a) Low-light &\hspace{-4mm} (b) Scene intensity &\hspace{-4mm} (c) Gated image
\\
\hspace{-1.5mm}\includegraphics[width = 0.325\linewidth]{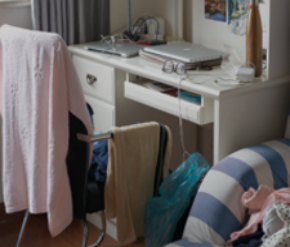} &\hspace{-4.5mm}
\includegraphics[width = 0.325\linewidth]{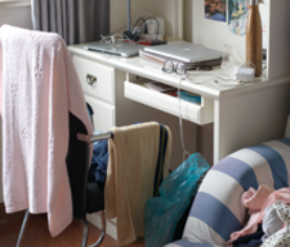} &\hspace{-4.5mm}
\includegraphics[width = 0.325\linewidth]{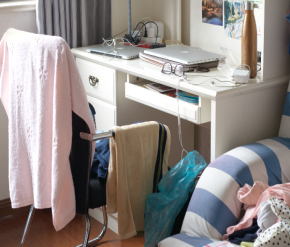} 
\\
\hspace{-1.5mm}(d)  w/o intensity &\hspace{-4mm} (e) \textbf{Ours} &\hspace{-4mm} (f) GT
\end{tabular}
\end{center}
\vspace{-2mm}
\caption{Effect of scene intensity. Using scene intensity to gate the low-light image as input significantly improves the results.
}
\label{fig:p6}
\end{figure}

%

To evaluate the robustness of our method under challenging real-world degradations, we test it on real-world datasets suffering from motion blur and noise respectively.
Fig.~\ref{fig:xin1} provides visual results on blurred low-light images from Real-LOL-Blur~\cite{LOLBLUR}.
MambaLIE generates much sharper images with visually pleasing results.
Fig.~\ref{fig:xin2} presents visual results on noisy low-light images from SIDD~\cite{SIDD}. 
MambaLIE produces clean images
while preserving fine textures.
Table~\ref{tab:xin1} presents the quantitative evaluation using NIQE and MUSIQ metrics. MambaLIE achieves the lowest NIQE score, indicating better perceptual quality, and the highest MUSIQ score, reflecting improved visual quality from an aesthetic perspective
These results collectively demonstrate the effectiveness and robustness of MambaLIE under real-world degradations in low-light scenarios.

\subsection{Ablation Study}
We conduct ablation studies to measure the contributions of the following factors: (1) Scene Light Intensity, and (2) Locally Enhanced State Space Model.
The base model is based on the approach proposed in resblock \cite{Bee2017resblock}.
\\
\noindent \textbf{Effect on Scene Light Intensity Prior.} 
As discussed in Section III-B, the scene light intensity prior is introduced to spatially modulate the low-light input and guide the enhancement process. To assess its effectiveness, we conduct an ablation study by removing the prior from the input pathway. 
\begin{table}[!t]
\centering
\scriptsize 
\tablestyle{1pt}{1}
\caption{Ablation study of the the prior filter size.}
\renewcommand{\arraystretch}{1.5}
\begin{tabular}{c|c|ccc|c|c}
\shline
Exp.  & Prior filter size & PSNR$\uparrow$& SSIM$\uparrow$& LPIPS$\downarrow$&Parameters &Inference time  \\
\shline
(a)  &$1 \times 1$&22.23& 0.723& 0.187 &13.5M    &47ms\\
(b) & $3 \times 3$  &23.52 &0.812& 0.163   &13.8M   &48ms\\
\textbf{(c)} & $\mathbf{5\times 5}$ & \textbf{23.86} & \textbf{0.826} & \textbf{0.153} & \textbf{14M} & \textbf{50ms} \\
(d) & $7 \times 7$&23.85& 0.823& 0.155   &14.3 M  &52ms\\
\shline
\end{tabular}
\label{tab:xin3}
\end{table}
Table~\ref{tab:t4} shows that simply employing the scene light intensity prior significantly improves the enhancement results, yielding a 3.1 dB gain in terms of PSNR, indicating improved global illumination estimation and reconstruction quality.
Fig.~\ref{fig:p6} further presents a visual comparison, showing that the scene intensity prior helps recover clearer structural content and more vivid color distribution. 
These results empirically validate the theoretical design, confirming that the light intensity prior plays a critical role in enhancing both local details and overall scene visibility.
Table~\ref{tab:xin3} reports the effects of different prior filter sizes, which determine the spatial scale for estimating scene light intensity. The configuration with a $5 \times 5$ convolutional filter achieves a favorable balance between enhancement quality and computational efficiency, and is thus adopted as the default setting.
\begin{figure}[t]\footnotesize
    \centering
    \begin{minipage}[t]{0.245\linewidth}
        \centering
        \includegraphics[width=\textwidth]{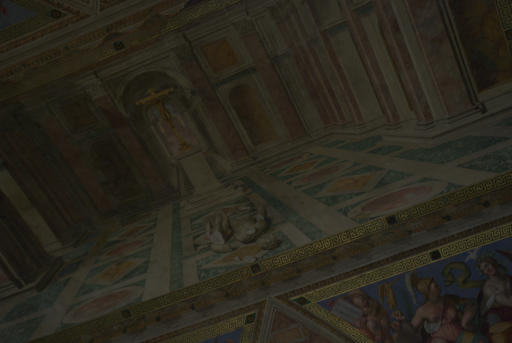}
        \subcaption*{(a) Low-light}
    \end{minipage}\hspace{-0.5mm}
    \begin{minipage}[t]{0.245\linewidth}
        \centering
        \includegraphics[width=\textwidth]{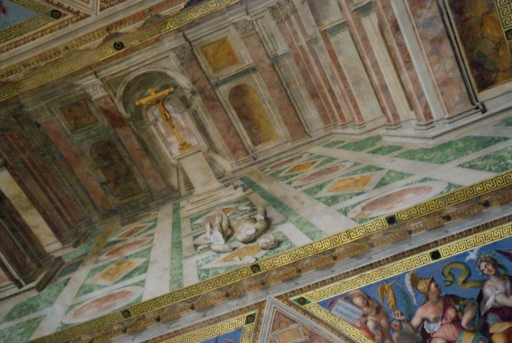}
        \subcaption*{(b) Trans. attn.}
    \end{minipage}\hspace{-0.5mm}
    \begin{minipage}[t]{0.245\linewidth}
        \centering
        \includegraphics[width=\textwidth]{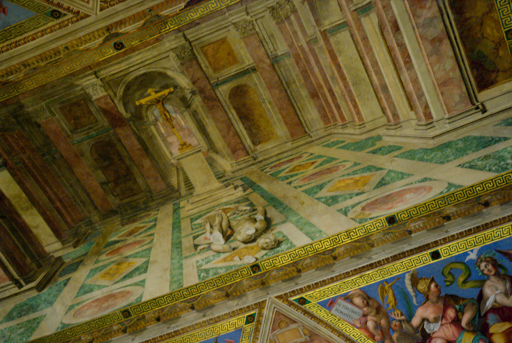}
        \subcaption*{\textbf{(c)} \textbf{Ours}}
    \end{minipage}\hspace{-0.5mm}
    \begin{minipage}[t]{0.245\linewidth}
        \centering
        \includegraphics[width=\textwidth]{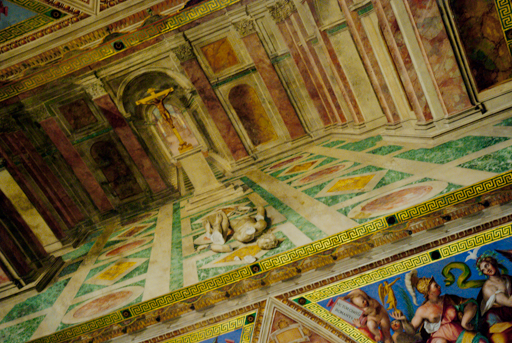}
        \subcaption*{(d) GT}
    \end{minipage}
    \caption{LESSM vs. Transposed attention (Trans. attn.). 
    Replacing Transposed attention~\cite{cvpr/ZamirA0HK022} with LESSM, the results are significantly improved.}
    \label{fig:p7}
    \vspace{-3mm}
\end{figure}

\noindent \textbf{Effect on Locally Enhanced State Space Model.} 
As we use LESSM to model long-range dependencies instead of regular attention such as transposed attention~\cite{cvpr/ZamirA0HK022}, it is crucial to measure the impact of the proposed LESSM.
%
Table~\ref{tab:t4} shows that using LESSM to replace regular attention significantly improves performance,  
resulting in a $2.73$ dB PSNR gains in PSNR.
Fig.~\ref{fig:p7} further demonstrates that our proposed LESSM significantly outperforms previous self-attention mechanisms~\cite{cvpr/ZamirA0HK022} when replacing self-attention with LESSM in our model.
%


\noindent\textbf{Effect on Channel Atention and Dual Gated Feed-Forward Network (DGFN).}
\begin{table}[!t]
\vspace{1.5mm}
\centering
\caption{Ablation study. 
Both scene intensity gated with low-light images and LESSM replaced by regular attention are effective in helping enhancement.}
\tablestyle{4pt}{1}
\scriptsize 
\renewcommand{\arraystretch}{1.25}
\begin{tabular}{c|cccc|ccc}
\shline
Exp.  & Base & Trans. Attn. & Scene intensity & LESSM & PSNR& SSIM& LPIPS \\
\shline
(a)  & \CheckmarkBold & \XSolidBrush & \XSolidBrush & \XSolidBrush & 21.55 &0.875&0.085\\
(b) & \CheckmarkBold  & \CheckmarkBold & \CheckmarkBold & \XSolidBrush &22.73&0.918&0.056\\
(c) & \CheckmarkBold  & \XSolidBrush & \XSolidBrush & \CheckmarkBold & 24.28 &0.925&0.044\\
\shline
(d) & \CheckmarkBold  & \XSolidBrush & \CheckmarkBold & \CheckmarkBold &  \textbf{25.83}&\textbf{0.928}&\textbf{0.042} \\
\shline
\end{tabular}
\label{tab:t4}
\end{table}
\begin{table}[t]
\centering
\vspace{1.5mm}
\caption{Ablation study of the Channel Attention and DGFN.}
\tablestyle{11pt}{1}
\renewcommand{\arraystretch}{1.25}
\begin{tabular}{c|cccc|c}
\shline
Exp.  & Base & LESSM & CA & DGFN & PSNR \\
\shline
(a)  & \CheckmarkBold & \XSolidBrush & \XSolidBrush & \XSolidBrush  &21.55\\
(b) & \CheckmarkBold  & \CheckmarkBold & \XSolidBrush & \XSolidBrush &24.78\\
(c) & \CheckmarkBold  & \CheckmarkBold & \CheckmarkBold & \XSolidBrush  &24.94\\
(d) & \CheckmarkBold  & \CheckmarkBold & \XSolidBrush & \CheckmarkBold &25.46\\
\shline
(e) & \CheckmarkBold  & \CheckmarkBold & \CheckmarkBold & \CheckmarkBold &\textbf{25.76}\\
\shline
\end{tabular}
\label{tab:t2.5}
\end{table}
We measure the impact of the Channel Attention (CA)~\cite{cvpr/HuSS18} and DGFN \cite{aaai/WangZSLSL23}, as shown in Table~\ref{tab:t2.5}. We test the PSNR on the MIT-Adobe FiveK \cite{vladimir2011five5k}.
LESSM's ablation studies are in the Ablation Studies section of the main paper.
When Channel Attention is added to model (b) to form model (c), the PSNR increases by $0.16$db.
This proves the effectiveness of Channel Attention in enhancing channel modeling capability and reducing channel redundancy caused by linear structure of LESSM.
When DGFN is added to model (b) to form model (d), the PSNR increases by $0.68$db.
By flexibly adjusting FFN's structure and parameters, the LESSM effectively improves feature extraction and information integration, thereby enhancing performance and generalization on complex tasks.
When Channel Attention and DGFN are used (e), the PSNR reaches a maximum of $25.76$db. 
This demonstrates the effectiveness of Channel Attention and DGFN in
enhancing LESSM performance.

\noindent\textbf{Effect of the number of GLEM blocks.}
Table~\ref{tab:xin2} reports the effects of varying the number of GLEM blocks across the four encoder-decoder stages.
We evaluate four allocation strategies: linear $\{2,4,6,8\}$, exponential $\{1,2,4,8\}$, saturated $\{4,6,6,8\}$, and superlinear $\{2,4,8,12\}$, where more blocks are assigned to deeper layers to enhance semantic abstraction. 
Among these, the exponential strategy with $\{2,4,6,8\}$ achieves the best balance between performance and inference cost, and is therefore selected as the default.

\begin{table}[!t]
\centering
\scriptsize 
\tablestyle{1pt}{1}
\caption{Ablation study of the the number of GLEM blocks.}
\renewcommand{\arraystretch}{1.5}
\begin{tabular}{c|c|ccc|c|c}
\shline
Exp.  & GLEM & PSNR$\uparrow$& SSIM$\uparrow$& LPIPS$\downarrow$&Parameters &Inference time  \\
\shline
(a)  &$\{1, 1, 1, 2\}$   &22.08& 0.753& 0.185 &4M    &16ms\\
\textbf{(b)} & $\mathbf{\{1,2,4,8\}}$ & \textbf{23.86} & \textbf{0.826} & \textbf{0.153} & \textbf{14M} & \textbf{50ms} \\
(c) & $\{2, 4, 6, 8\}$  &23.46& 0.805& 0.161   &17M   &67ms\\
(d) & $\{4, 6, 6, 8\}$    &23.77& 0.821& 0.157   &22M   &80ms\\
(e) & $\{2, 4, 8, 12\}$   &23.85& 0.828& 0.151    &24M    &87ms\\
\shline
\end{tabular}
\label{tab:xin2}
\end{table}

\begin{table}[!t]
\centering
\scriptsize 
\caption{Ablation study of the Local Enhanced Branch and SSM Branch.}
\tablestyle{1pt}{1}
\renewcommand{\arraystretch}{1.5}
\begin{tabular}{c|ccc|c}
\shline
Exp.  & Base Model & SSM Branch& Local Enhanced Branch&PSNR  \\
\shline
(a)  &\CheckmarkBold   &\XSolidBrush   & \XSolidBrush& 21.55 \\
(b) & \CheckmarkBold  &\XSolidBrush & \CheckmarkBold & 21.85   \\
(c) & \CheckmarkBold     &\CheckmarkBold & \XSolidBrush  & 23.52   \\

\shline
(d) & \CheckmarkBold   &\CheckmarkBold &\CheckmarkBold &23.86 \\
\shline
\end{tabular}
\label{tab:txin5}
\end{table}

\begin{table}[!t]
\centering
\scriptsize 
\caption{Ablation study of the ForwardSSM and BackwardSSM.}
\renewcommand{\arraystretch}{1.5}
\begin{tabular}{c|ccc|c}
\shline
Exp.  &Base Model& ForwardSSM & BackwardSSM&PSNR  \\
\shline
(a) &\CheckmarkBold & \XSolidBrush    &\XSolidBrush   & 21.25   \\
(b)  &\CheckmarkBold&\CheckmarkBold   &\XSolidBrush  & 22.34 \\
(c) &\CheckmarkBold& \XSolidBrush  &\CheckmarkBold & 22.18   \\
\shline
(d) &\CheckmarkBold& \CheckmarkBold   &\CheckmarkBold  &23.86 \\
\shline
\end{tabular}
\label{tab:txin4}
\vspace{-4mm}
\end{table}

\noindent\textbf{Effect of the Local Enhanced Branch and SSM Branch.}
We explore the trade-off between the Local Enhanced Branch and SSM Branch. As shown in Table~\ref{tab:txin5}, the results reveal that adding either the SSM Branch or the Local Enhanced Branch improves PSNR, with the SSM Branch alone providing a larger gain. When combined, the two branches enable the model to achieve its best overall performance.

\noindent\textbf{Effect of Different SSM Configurations.}
We investigate the impact of various SSM configurations in MambaLIE.
Table~\ref{tab:txin4} shows that the bidirectional SSM achieves the best results by capturing information from both forward and backward directions. 
This allows the model to better understand the context in low-light scenes, leading to improved restoration performance.

\begin{figure}[!t]
\includegraphics[width=1\linewidth]{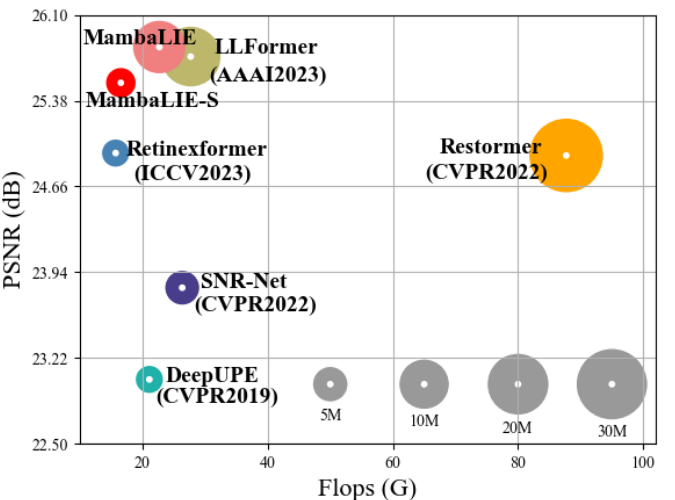}

\caption{Model Flops comparison on MIT-Adobe FiveK. MambaLIE achieves a better trade-off between model complexity and performance. The number of encoder ande decoder blocks in the MambaLIE model is reduced from $\{1, 2, 4, 8\}$ in stages 1 to 4 to $\{1, 1, 1, 2\}$ to obtain MambaLIE-S.}
\label{fig:p8}
\vspace{-2mm}
\end{figure}

\noindent\textbf{Model Complexity Comparisons}
%
%
We further compare the model complexity in terms of Flops and model size in Fig.~\ref{fig:p8}.
%
%
Our MambaLIE strikes a better trade-off balance between model size, performance, and computational complexity as measured by Flops. 
Specifically, our MambaLIE ranks first in performance on the MIT-Adobe FiveK dataset while boasting fewer model parameters and lower Flops.
We propose a smaller model version, MambaLIE-S, which achieves a higher PSNR index with reduced parameters and Flops.

\subsection{Limitations and Future Work}
%
%
This study involves the utilization of the scene light intensity prior in the MambaLIE model to better capture structural information and enhance image details effectively. 
Although the MambaLIE model has made significant progress in low-light image enhancement, it still has some limitations. 
%
%

\begin{figure}[t]\footnotesize
    \centering
    \begin{minipage}[t]{0.45\linewidth}
        \centering
        \includegraphics[width=\textwidth]{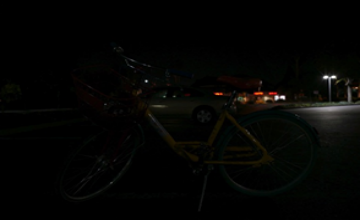}
        \subcaption*{(a) Input}
    \end{minipage}\hspace{-0.5mm}
    \begin{minipage}[t]{0.45\linewidth}
        \centering
        \includegraphics[width=\textwidth]{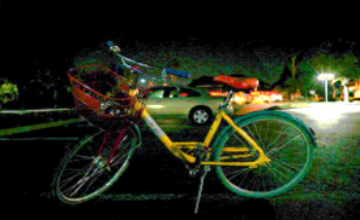}
        \subcaption*{(b) Ours}
    \end{minipage}\hspace{-0.5mm}
    \vspace{-1mm}
\caption{Limited enhancement quality under extreme low-light conditions. MambaLIE enhances overall visibility, it still exhibits color distortion as well as under-/over-exposed  in severrly degarded regions.}
\label{fig:pxin3}
    \vspace{-2mm}
\end{figure}

%
In extremely complex lighting conditions, the model may struggle to restore fine details, leading to color distortion and exposure artifacts. As shown in Fig.~\ref{fig:pxin3}, MambaLIE exhibits color shifts, nder-/over-exposed in severely degraded regions. These issues stem from inaccurate prior estimation in dark areas and the sensitivity of global modeling to non-uniform illumination.
Future work will focus on adaptive illumination priors and robust global modeling mechanisms to handle extreme lighting variations, along with lightweight model designs for efficient deployment.

%


\subsection{Application in Consumer Electronics}
\begin{figure}[t]\footnotesize
    \centering
    \begin{minipage}[t]{0.45\linewidth}
        \centering
        \includegraphics[width=\textwidth]{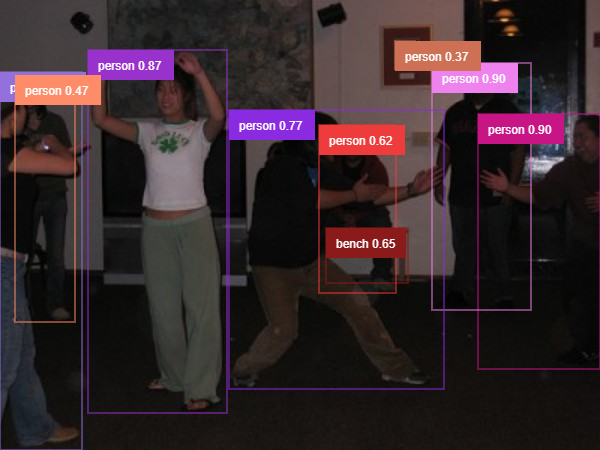}
        \subcaption*{(a)}
    \end{minipage}\hspace{-0.5mm}
    \begin{minipage}[t]{0.45\linewidth}
        \centering
        \includegraphics[width=\textwidth]{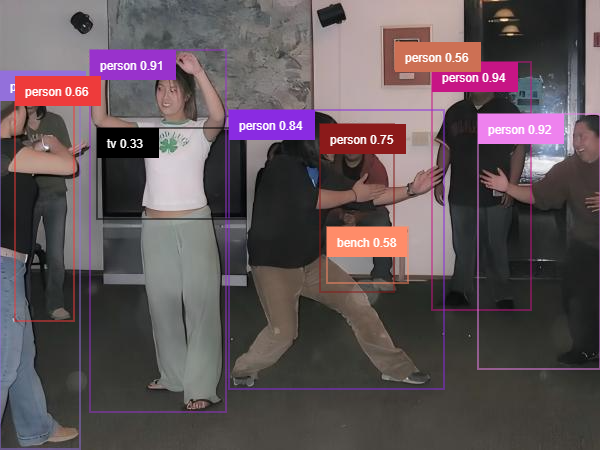}
        \subcaption*{(b)}
    \end{minipage}\hspace{-0.5mm}
    \vspace{-1mm}
\caption{Visual comparison of object detection in consumer electronic devices before and after low-light image enhancement. 
(a) Detection result on the low-light image captured under challenging illumination conditions. 
(b) Detection result on the enhanced image, demonstrating improved illumination and more reliable detection performance.}
\label{fig:pxin11}
    \vspace{-2mm}
\end{figure}

In consumer electronic imaging systems, low-light conditions degrade visual quality and reduce the reliability of downstream vision modules. 
As shown in Fig.~\ref{fig:pxin11}, the proposed method enhances illumination consistency and structural details, leading to improved object detection performance. 
Specifically, objects that are difficult to recognize in low-light inputs become more distinguishable after enhancement, resulting in more accurate localization and classification. 
This demonstrates the practical value of the proposed method in consumer electronic pipelines, such as mobile imaging, intelligent surveillance, and edge-based vision systems.

%

%
\section{Conclusion}
We have proposed MambaLIE, a scene light intensity-boosted LIE with a state space model. The scene light intensity is used to perceive the light intensity and further reveal useful structures to help enhance images.  We further propose the locally enhanced state space model to explore long-range dependencies and enhance feature representation while maintaining linear computation efficiency for efficient and effective image enhancement.
Experiments have shown that our MambaLIE performs exceptionally well compared to state-of-the-art CNN-based and Transformers-based LIE methods on four synthetic datasets and five real-world benchmarks, demonstrating its effectiveness and efficiency for practical low-light imaging and its suitability for deployment on resource-constrained consumer electronic devices.

%



%


\bibliographystyle{IEEEtran}
\bibliography{sample.bib}

@article{dsp/ChengS04,
  author       = {Heng{-}Da Cheng and
                  X. J. Shi},
  title        = {A simple and effective histogram equalization approach to image enhancement},
  journal      = {DSP},
  volume       = {14},
  number       = {2},
  pages        = {158--170},
  year         = {2004}
}

@article{tip/LeeLK13,
  author       = {Chulwoo Lee and
                  Chul Lee and
                  Chang{-}Su Kim},
  title        = {Contrast Enhancement Based on Layered Difference Representation of
                  2D Histograms},
  journal      = {{IEEE} TIP},
  volume       = {22},
  number       = {12},
  pages        = {5372--5384},
  year         = {2013}
}

@article{displays/WangLL09,
  author       = {Zhiguo Wang and
                  Zhi{-}Hu Liang and
                  Chun{-}Liang Liu},
  title        = {A real-time image processor with combining dynamic contrast ratio
                  enhancement and inverse gamma correction for {PDP}},
  journal      = {Displays},
  volume       = {30},
  number       = {3},
  pages        = {133--139},
  year         = {2009}
}

@article{ejivp/RahmanRAAS16,
  author       = {Shanto Rahman and
                  Mostafijur Rahman and
                  Mohammad Abdullah{-}Al{-}Wadud and
                  Golam Dastegir Al{-}Quaderi and
                  Mohammad Shoyaib},
  title        = {An adaptive gamma correction for image enhancement},
  journal      = {EJIVP},
  volume       = {2016},
  pages        = {35},
  year         = {2016}
}

@inproceedings{bmvc/WeiWY018,
  author       = {Chen Wei and
                  Wenjing Wang and
                  Wenhan Yang and
                  Jiaying Liu},
  title        = {Deep Retinex Decomposition for Low-Light Enhancement},
  booktitle    = {BMVC},
  pages        = {155},
  year         = {2018}
}

@inproceedings{mm/ZhangZG19,
  author       = {Yonghua Zhang and
                  Jiawan Zhang and
                  Xiaojie Guo},
  title        = {Kindling the Darkness: {A} Practical Low-light Image Enhancer},
  booktitle    = {{ACM} MM},
  pages        = {1632--1640},
  year         = {2019}
}

@inproceedings{cvpr/GuoLGLHKC20,
  author       = {Chunle Guo and
                  Chongyi Li and
                  Jichang Guo and
                  Chen Change Loy and
                  Junhui Hou and
                  Sam Kwong and
                  Runmin Cong},
  title        = {Zero-Reference Deep Curve Estimation for Low-Light Image Enhancement},
  booktitle    = {CVPR},
  pages        = {1777--1786},
  year         = {2020},
}

@inproceedings{cvpr/WuWZWYJ22,
  author       = {Wenhui Wu and
                  Jian Weng and
                  Pingping Zhang and
                  Xu Wang and
                  Wenhan Yang and
                  Jianmin Jiang},
  title        = {URetinex-Net: Retinex-based Deep Unfolding Network for Low-light Image
                  Enhancement},
  booktitle    = {CVPR},
  pages        = {5891--5900},
  year         = {2022}
}

@inproceedings{cvpr/XuWFJ22,
  author       = {Xiaogang Xu and
                  Ruixing Wang and
                  Chi{-}Wing Fu and
                  Jiaya Jia},
  title        = {SNR-Aware Low-light Image Enhancement},
  booktitle    = {CVPR},
  pages        = {17693--17703},
  year         = {2022}
}

@inproceedings{iccv/CaiBLWTZ23,
  author       = {Yuanhao Cai and
                  Hao Bian and
                  Jing Lin and
                  Haoqian Wang and
                  Radu Timofte and
                  Yulun Zhang},
  title        = {Retinexformer: One-stage Retinex-based Transformer for Low-light Image
                  Enhancement},
  booktitle    = {ICCV},
  pages        = {12470--12479},
  year         = {2023}
}

@inproceedings{cvpr/ZamirA0HK022,
  author       = {Syed Waqas Zamir and
                  Aditya Arora and
                  Salman Khan and
                  Munawar Hayat and
                  Fahad Shahbaz Khan and
                  Ming{-}Hsuan Yang},
  title        = {Restormer: Efficient Transformer for High-Resolution Image Restoration},
  booktitle    = {CVPR},
  pages        = {5718--5729},
  publisher    = {{IEEE}},
  year         = {2022}
}

@article{corr/abs-2312-00752,
  author       = {Albert Gu and
                  Tri Dao},
  title        = {Mamba: Linear-Time Sequence Modeling with Selective State Spaces},
  journal      = {arXiv:2312.00752},
  year         = {2023}
}

@inproceedings{vmamba,
  author       = {Yue Liu and
                  Yunjie Tian and
                  Yuzhong Zhao and
                  Hongtian Yu and
                  Lingxi Xie and
                  Yaowei Wang and
                  Qixiang Ye and
                  Jianbin Jiao and
                  Yunfan Liu},
  title        = {VMamba: Visual State Space Model},
  booktitle    = {NeurIPS},
  year         = {2024},
}

@article{tip/WangZHL13,
  author       = {Shuhang Wang and
                  Jin Zheng and
                  Hai{-}Miao Hu and
                  Bo Li},
  title        = {Naturalness Preserved Enhancement Algorithm for Non-Uniform Illumination
                  Images},
  journal      = {{IEEE} TIP},
  volume       = {22},
  number       = {9},
  pages        = {3538--3548},
  year         = {2013}
}

@inproceedings{cvpr/FuZHZD16,
  author       = {Xueyang Fu and
                  Delu Zeng and
                  Yue Huang and
                  Xiao{-}Ping (Steven) Zhang and
                  Xinghao Ding},
  title        = {A Weighted Variational Model for Simultaneous Reflectance and Illumination
                  Estimation},
  booktitle    = {CVPR},
  pages        = {2782--2790},
  year         = {2016}
}

@inproceedings{nips/GuG0R22,
  author       = {Albert Gu and
                  Karan Goel and
                  Ankit Gupta and
                  Christopher R{\'{e}}},
  title        = {On the Parameterization and Initialization of Diagonal State Space
                  Models},
  booktitle    = {NeurIPS},
  year         = {2022}
}

@inproceedings{iclr/GuGR22,
  author       = {Albert Gu and
                  Karan Goel and
                  Christopher R{\'{e}}},
  title        = {Efficiently Modeling Long Sequences with Structured State Spaces},
  booktitle    = {ICLR},
  year         = {2022}
}

@inproceedings{vim,
  author       = {Lianghui Zhu and
                  Bencheng Liao and
                  Qian Zhang and
                  Xinlong Wang and
                  Wenyu Liu and
                  Xinggang Wang},
  title        = {Vision Mamba: Efficient Visual Representation Learning with Bidirectional
                  State Space Model},
  booktitle    = {ICML},
  volume       = {235},
  pages        = {62429--62442},
  year         = {2024},
}

@article{sigpro/FuZHLDP16,
  author       = {Xueyang Fu and
                  Delu Zeng and
                  Yue Huang and
                  Yinghao Liao and
                  Xinghao Ding and
                  John W. Paisley},
  title        = {A fusion-based enhancing method for weakly illuminated images},
  journal      = {Signal Process.},
  volume       = {129},
  pages        = {82--96},
  year         = {2016}
}

@article{pami/LiGL22,
  author       = {Chongyi Li and
                  Chunle Guo and
                  Chen Change Loy},
  title        = {Learning to Enhance Low-Light Image via Zero-Reference Deep Curve
                  Estimation},
  journal      = {{IEEE} TPAMI},
  volume       = {44},
  number       = {8},
  pages        = {4225--4238},
  year         = {2022}
}

@article{ijcv/ZhangGMLZ21,
  author       = {Yonghua Zhang and
                  Xiaojie Guo and
                  Jiayi Ma and
                  Wei Liu and
                  Jiawan Zhang},
  title        = {Beyond Brightening Low-light Images},
  journal      = {IJCV},
  volume       = {129},
  number       = {4},
  pages        = {1013--1037},
  year         = {2021}
}

@article{tip/JiangGLCFSYZW21,
  author       = {Yifan Jiang and
                  Xinyu Gong and
                  Ding Liu and
                  Yu Cheng and
                  Chen Fang and
                  Xiaohui Shen and
                  Jianchao Yang and
                  Pan Zhou and
                  Zhangyang Wang},
  title        = {EnlightenGAN: Deep Light Enhancement Without Paired Supervision},
  journal      = {{IEEE} TIP},
  volume       = {30},
  pages        = {2340--2349},
  year         = {2021}
}

@inproceedings{cvpr/Liu0Z0L21,
  author       = {Risheng Liu and
                  Long Ma and
                  Jiaao Zhang and
                  Xin Fan and
                  Zhongxuan Luo},
  title        = {Retinex-Inspired Unrolling With Cooperative Prior Architecture Search
                  for Low-Light Image Enhancement},
  booktitle    = {CVPR},
  pages        = {10561--10570},
  year         = {2021}
}

@inproceedings{cvpr/WangCBZLL22,
  author       = {Zhendong Wang and
                  Xiaodong Cun and
                  Jianmin Bao and
                  Wengang Zhou and
                  Jianzhuang Liu and
                  Houqiang Li},
  title        = {Uformer: {A} General U-Shaped Transformer for Image Restoration},
  booktitle    = {CVPR},
  pages        = {17662--17672},
  year         = {2022}
}

@inproceedings{mm/WangWJ23,
  author       = {Chenxi Wang and
                  Hongjun Wu and
                  Zhi Jin},
  title        = {FourLLIE: Boosting Low-Light Image Enhancement by Fourier Frequency
                  Information},
  booktitle    = {{ACM} {MM}},
  pages        = {7459--7469},
  year         = {2023}
}

@inproceedings{aaai/WangZSLSL23,
  author       = {Tao Wang and
                  Kaihao Zhang and
                  Tianrun Shen and
                  Wenhan Luo and
                  Bj{\"{o}}rn Stenger and
                  Tong Lu},
  title        = {Ultra-High-Definition Low-Light Image Enhancement: {A} Benchmark and
                  Transformer-Based Method},
  booktitle    = {AAAI},
  pages        = {2654--2662},
  year         = {2023}
}

@article{corr/abs-2406-01028,
  author       = {Xuanqi Zhang and
                  Haijin Zeng and
                  Jinwang Pan and
                  Qiangqiang Shen and
                  Yongyong Chen},
  title        = {LLEMamba: Low-Light Enhancement via Relighting-Guided Mamba with Deep
                  Unfolding Network},
  journal      = {arXiv:2406.01028},
  year         = {2024}
}

@article{ijcv/GuoH23,
  author       = {Xiaojie Guo and
                  Qiming Hu},
  title        = {Low-light Image Enhancement via Breaking Down the Darkness},
  journal      = {IJCV},
  volume       = {131},
  number       = {1},
  pages        = {48--66},
  year         = {2023}
}

@article{dsp/WeiLL24,
  author       = {Xinxu Wei and
                  Xi Lin and
                  Yongjie Li},
  title        = {{DA-DRN:} {A} degradation-aware deep Retinex network for low-light
                  image enhancement},
  journal      = {DSP},
  volume       = {144},
  pages        = {104256},
  year         = {2024}
}

@article{tcsv/LuoYYL24,
  author       = {Yu Luo and
                  Bijia You and
                  Guanghui Yue and
                  Jie Ling},
  title        = {Pseudo-Supervised Low-Light Image Enhancement With Mutual Learning},
  journal      = {{IEEE} TCSVT},
  volume       = {34},
  number       = {1},
  pages        = {85--96},
  year         = {2024}
}

@inproceedings{Jiang2024LightenDiffusionUL,
  title={LightenDiffusion: Unsupervised Low-Light Image Enhancement with Latent-Retinex Diffusion Models},
  author={Hai Jiang and Ao Luo and Xiaohong Liu and Songchen Han and Shuaicheng Liu},
  booktitle= {ECCV},
  year={2024}
}

@article{tip/GuoLL17,
  author       = {Xiaojie Guo and
                  Yu Li and
                  Haibin Ling},
  title        = {{LIME:} Low-Light Image Enhancement via Illumination Map Estimation},
  journal      = {{IEEE} TIP},
  volume       = {26},
  number       = {2},
  pages        = {982--993},
  year         = {2017}
}

@inproceedings{iccv/YangDWLZ23,
  author       = {Shuzhou Yang and
                  Moxuan Ding and
                  Yanmin Wu and
                  Zihan Li and
                  Jian Zhang},
  title        = {Implicit Neural Representation for Cooperative Low-light Image Enhancement},
  booktitle    = {ICCV},
  pages        = {12872--12881},
  year         = {2023}
}

@article{cviu/DangZQ24,
  author       = {Jiachen Dang and
                  Yong Zhong and
                  Xiaolin Qin},
  title        = {PPformer: Using pixel-wise and patch-wise cross-attention for low-light
                  image enhancement},
  journal      = {CVIU},
  volume       = {241},
  pages        = {103930},
  year         = {2024}
}

@article{cviu/LiWFL24,
  author       = {Xiaofang Li and
                  Weiwei Wang and
                  Xiangchu Feng and
                  Min Li},
  title        = {Deep parametric Retinex decomposition model for low-light image enhancement},
  journal      = {CVIU},
  volume       = {241},
  pages        = {103948},
  year         = {2024}
}

@article{wenhua2021lolv2,
  author       = {Wenhan Yang and
                  Wenjing Wang and
                  Haofeng Huang and
                  Shiqi Wang and
                  Jiaying Liu},
  title        = {Sparse Gradient Regularized Deep Retinex Network for Robust Low-Light
                  Image Enhancement},
  journal      = {{IEEE} TIP},
  volume       = {30},
  pages        = {2072--2086},
  year         = {2021}
}

@inproceedings{vladimir2011five5k,
  author       = {Vladimir Bychkovsky and
                  Sylvain Paris and
                  Eric Chan and
                  Fr{\'{e}}do Durand},
  title        = {Learning photographic global tonal adjustment with a database of input
                  / output image pairs},
  booktitle    = {CVPR},
  pages        = {97--104},
  year         = {2011}
}

@article{kede2015MEF,
  author       = {Kede Ma and
                  Kai Zeng and
                  Zhou Wang},
  title        = {Perceptual Quality Assessment for Multi-Exposure Image Fusion},
  journal      = {{IEEE} TIP},
  volume       = {24},
  number       = {11},
  pages        = {3345--3356},
  year         = {2015}
}

@inproceedings{Bee2017resblock,
  author       = {Bee Lim and
                  Sanghyun Son and
                  Heewon Kim and
                  Seungjun Nah and
                  Kyoung Mu Lee},
  title        = {Enhanced Deep Residual Networks for Single Image Super-Resolution},
  booktitle    = {CVPR},
  pages        = {1132--1140},
  year         = {2017}
}

@article{PSNR_thu,
  author={Huynh-Thu, Q. and Ghanbari, M.},
  title={Scope of validity of PSNR in image/video quality assessment},
  journal={Electronics Letters},
  volume={44},
  number={13},
  pages={800-801},
  year={2008},
}

@article{SSIM_wang,
  author    = {Zhou Wang and
               Alan C. Bovik and
               Hamid R. Sheikh and
               Eero P. Simoncelli},
  title     = {Image quality assessment: from error visibility to structural similarity},
  journal   = {{IEEE} TIP},
  volume    = {13},
  number    = {4},
  pages     = {600--612},
  year      = {2004},
}

@inproceedings{LPIPS,
  author       = {Richard Zhang and
                  Phillip Isola and
                  Alexei A. Efros and
                  Eli Shechtman and
                  Oliver Wang},
  title        = {The Unreasonable Effectiveness of Deep Features as a Perceptual Metric},
  booktitle    = {CVPR},
  pages        = {586--595},
  year         = {2018}
}

@article{niqe,
  author       = {Anish Mittal and
                  Rajiv Soundararajan and
                  Alan C. Bovik},
  title        = {Making a "Completely Blind" Image Quality Analyzer},
  journal      = {{IEEE} SPL},
  volume       = {20},
  number       = {3},
  pages        = {209--212},
  year         = {2013}
}

@article{pi,
   author       = {Chao Ma and
                  Chih{-}Yuan Yang and
                  Xiaokang Yang and
                  Ming{-}Hsuan Yang},
  title        = {Learning a no-reference quality metric for single-image super-resolution},
  journal      = {CVIU},
  volume       = {158},
  pages        = {1--16},
  year         = {2017}
}

@inproceedings{adam,
   author       = {Diederik P. Kingma and
                  Jimmy Ba},
  title        = {Adam: {A} Method for Stochastic Optimization},
  booktitle    = {ICLR},
  year         = {2015}
}

@inproceedings{loshchilov2016sgdr,
  author       = {Ilya Loshchilov and
                  Frank Hutter},
  title        = {{SGDR:} Stochastic Gradient Descent with Warm Restarts},
  booktitle    = {ICLR},
  year         = {2017},
}

@inproceedings{eccv/ZhangLLWZF18,
  author       = {Yulun Zhang and
                  Kunpeng Li and
                  Kai Li and
                  Lichen Wang and
                  Bineng Zhong and
                  Yun Fu},
  title        = {Image Super-Resolution Using Very Deep Residual Channel Attention
                  Networks},
  booktitle    = {ECCV},
  volume       = {11211},
  pages        = {294--310},
  year         = {2018},
}

@article{zou2023object,
  author       = {Zhengxia Zou and
                  Keyan Chen and
                  Zhenwei Shi and
                  Yuhong Guo and
                  Jieping Ye},
  title        = {Object Detection in 20 Years: {A} Survey},
  journal      = {Proc. {IEEE}},
  volume       = {111},
  number       = {3},
  pages        = {257--276},
  year         = {2023},
}

@inproceedings{li2023mseg3d,
  title={Mseg3d: Multi-modal 3d semantic segmentation for autonomous driving},
  author={Li, Jiale and Dai, Hang and Han, Hao and Ding, Yong},
  booktitle={CVPR},
  pages={21694--21704},
  year={2023}
}

@article{corr/BaKH16,
  author       = {Lei Jimmy Ba and
                  Jamie Ryan Kiros and
                  Geoffrey E. Hinton},
  title        = {Layer Normalization},
  journal      = {arXiv:1607.06450},
  year         = {2016},
}

@inproceedings{cvpr/HuSS18,
  author       = {Jie Hu and
                  Li Shen and
                  Gang Sun},
  title        = {Squeeze-and-Excitation Networks},
  booktitle    ={CVPR},
  pages        = {7132--7141},
  year         = {2018},
}

@inproceedings{nips/VaswaniSPUJGKP17,
  author       = {Ashish Vaswani and
                  Noam Shazeer and
                  Niki Parmar and
                  Jakob Uszkoreit and
                  Llion Jones and
                  Aidan N. Gomez and
                  Lukasz Kaiser and
                  Illia Polosukhin},
  title        = {Attention is All you Need},
  booktitle    = {NeurIPS},
  pages        = {5998--6008},
  year         = {2017},
}

@article{TciRIRO,
  author       = {Lijun Zhao and
                  Bintao Chen and
                  Jinjing Zhang and
                  Anhong Wang and
                  Huihui Bai},
  title        = {{RIRO:} From Retinex-Inspired Reconstruction Optimization Model to
                  Deep Low-Light Image Enhancement Unfolding Network},
  journal      = {{IEEE} TCI},
  volume       = {10},
  pages        = {969--983},
  year         = {2024},
}

@article{TCIDual-Stream,
  author       = {Guang Han and
                  Kang Wu and
                  Fanyu Zeng and
                  Jixin Liu and
                  Sam Kwong},
  title        = {Dual-Stream Adaptive Convergent Low-Light Image Enhancement Network
                  Based on Frequency Perception},
  journal      = {{IEEE} TCI},
  volume       = {9},
  pages        = {1152--1164},
  year         = {2023},
}

@ARTICLE{tciLFIENet,
  author={Ye, Wuyang and Yan, Tao and Gao, Jiahui and Yang, Yang},
  title={LFIENet: Light Field Image Enhancement Network by Fusing Exposures of LF-DSLR Image Pairs}, 
  journal      = {{IEEE} TCI},
volume={9},
  pages={620-635},
  year={2023},
  

}

@article{retinextip,
  author       = {Zunjin Zhao and
                  Bangshu Xiong and
                  Lei Wang and
                  Qiaofeng Ou and
                  Lei Yu and
                  Fa Kuang},
  title        = {RetinexDIP: {A} Unified Deep Framework for Low-Light Image Enhancement},
  journal      = {{IEEE} TCSVT},
  volume       = {32},
  number       = {3},
  pages        = {1076--1088},
  year         = {2022},
}

@article{EFINet,
  author       = {Chunxiao Liu and
                  Fanding Wu and
                  Xun Wang},
  title        = {EFINet: Restoration for Low-Light Images via Enhancement-Fusion Iterative
                  Network},
  journal      = {{IEEE} TCSVT},
  volume       = {32},
  number       = {12},
  pages        = {8486--8499},
  year         = {2022},
}

@article{swanet,
  author       = {Zhiquan He and
                  Wu Ran and
                  Shulin Liu and
                  Kehua Li and
                  Jiawen Lu and
                  Chang{-}Yong Xie and
                  Yong Liu and
                  Hong Lu},
  title        = {Low-Light Image Enhancement With Multi-Scale Attention and Frequency-Domain
                  Optimization},
  journal      = {{IEEE} TCSVT},
  volume       = {34},
  number       = {4},
  pages        = {2861--2875},
  year         = {2024},
}

@ARTICLE{cspn,
  author={Wu, Hongjun and Wang, Chenxi and Tu, Luwei and Patsch, Constantin and Jin, Zhi},
    
  title={CSPN: A Category-specific Processing Network for Low-light Image Enhancement}, 
journal      = {{IEEE} TCSVT},
volume={},
  number={},
  pages={1-1},
  year={2024},
  
}

@article{pwgcm,
  author       = {Xiangsheng Li and
                  Manlu Liu and
                  Qiang Ling},
  title        = {Pixel-Wise Gamma Correction Mapping for Low-Light Image Enhancement},
    journal      = {{IEEE} TCSVT},
  volume       = {34},
  number       = {2},
  pages        = {681--694},
  year         = {2024},
}

@ARTICLE{sclm,
   author={Zhang, Yuantong and Teng, Baoxin and Yang, Daiqin and Chen, Zhenzhong and Ma, Haichuan and Li, Gang and Ding, Wenpeng},
   
  title={Learning a Single Convolutional Layer Model for Low Light Image Enhancement}, 
 journal      = {{IEEE} TCSVT}, 
volume={34},
  number={7},
  pages={5995-6008},
  year={2024},
  
}

@article{shuixia1,
  author       = {Alireza Esmaeilzehi and
                  Yang Ou and
                  M. Omair Ahmad and
                  M. N. S. Swamy},
  title        = {{DMML:} Deep Multi-Prior and Multi-Discriminator Learning for Underwater
                  Image Enhancement},
  journal      = {{IEEE} Trans. Broadcast.},
  volume       = {70},
  number       = {2},
  pages        = {637--653},
  year         = {2024},
}

@article{chaofen1,
  author       = {Alireza Esmaeilzehi and
                  M. Omair Ahmad and
                  M. N. S. Swamy},
  title        = {{SRNMSM:} {A} Deep Light-Weight Image Super Resolution Network Using
                  Multi-Scale Spatial and Morphological Feature Generating Residual
                  Blocks},
  journal      = {{IEEE} Trans. Broadcast.},
  volume       = {68},
  number       = {1},
  pages        = {58--68},
  year         = {2022},
}

@article{chaofen2,
  author       = {Zhe Zhang and
                  Jianjun Lei and
                  Bo Peng and
                  Jie Zhu and
                  Qingming Huang},
  title        = {Self-Supervised Pretraining for Stereoscopic Image Super-Resolution
                  With Parallax-Aware Masking},
  journal      = {{IEEE} Trans. Broadcast.},
  volume       = {70},
  number       = {2},
  pages        = {482--491},
  year         = {2024},
}

@article{quwu1,
  author       = {Qiang Guo and
                  Mingliang Zhou},
  title        = {Progressive Domain Translation Defogging Network for Real-World Fog
                  Images},
  journal      = {{IEEE} Trans. Broadcast.},
  volume       = {68},
  number       = {4},
  pages        = {876--885},
  year         = {2022},
}

@inproceedings{SIDD,
  author       = {Abdelrahman Abdelhamed and
                  Stephen Lin and
                  Michael S. Brown},
  title        = {A High-Quality Denoising Dataset for Smartphone Cameras},
  booktitle    ={CVPR},
  pages        = {1692--1700},
  year         = {2018},
}

@inproceedings{LOLBLUR,
  author       = {Shangchen Zhou and
                  Chongyi Li and
                  Chen Change Loy},
  title        = {LEDNet: Joint Low-Light Enhancement and Deblurring in the Dark},
  booktitle    = {ECCV},
  volume       = {13666},
  pages        = {573--589},
  year         = {2022},
}

@inproceedings{s4nd,
  author       = {Eric Nguyen and
                  Karan Goel and
                  Albert Gu and
                  Gordon W. Downs and
                  Preey Shah and
                  Tri Dao and
                  Stephen Baccus and
                  Christopher R{\'{e}}},
  title        = {{S4ND:} Modeling Images and Videos as Multidimensional Signals with
                  State Spaces},
  booktitle    = {NeurIPS},
  year         = {2022}
}

@inproceedings{camera,
  author       = {Nan Qi and
                  Yeting Huang and
                  Wen Sun and
                  Shi Jin and
                  Theodoros A. Tsiftsis and
                  Qihui Wu and
                  Xiang Su},
  title        = {Unity makes strength: Coalition Formation-based Group-buying for Timely
                  {UAV} Data Collection},
  booktitle    = {GLOBECOM},
  pages        = {3712--3717},
  year         = {2022},
}

@ARTICLE{image,
  author={Kao, Wen-chung and Wang, Sheng-hong and Chen, Lien-yang and Lin, Sheng-yuan},
  journal={IEEE Transactions on Consumer Electronics}, 
  title={Design considerations of color image processing pipeline for digital cameras}, 
  year={2006},
  volume={52},
  number={4},
  pages={1144-1152},
  }

\vspace{-6mm}
\begin{IEEEbiography}[{\includegraphics[width=1in,height=1.25in,clip,keepaspectratio]{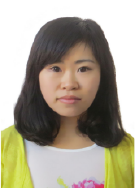}}]{Wanshu Fan} (Member, IEEE) was born in Heilongjiang, China. She received the Ph.D. degree in computational mathematics from the Dalian University of Technology in 2020. Currently, she is working in the School of Software Engineering of Dalian University. Her research interests include computer vision, artificial intelligence, machine learning and big data technology. She is a member of IEEE and CCF.
\end{IEEEbiography}

\vspace{-3mm}
\begin{IEEEbiography}[{\includegraphics[width=1in,height=1.25in,clip,keepaspectratio]{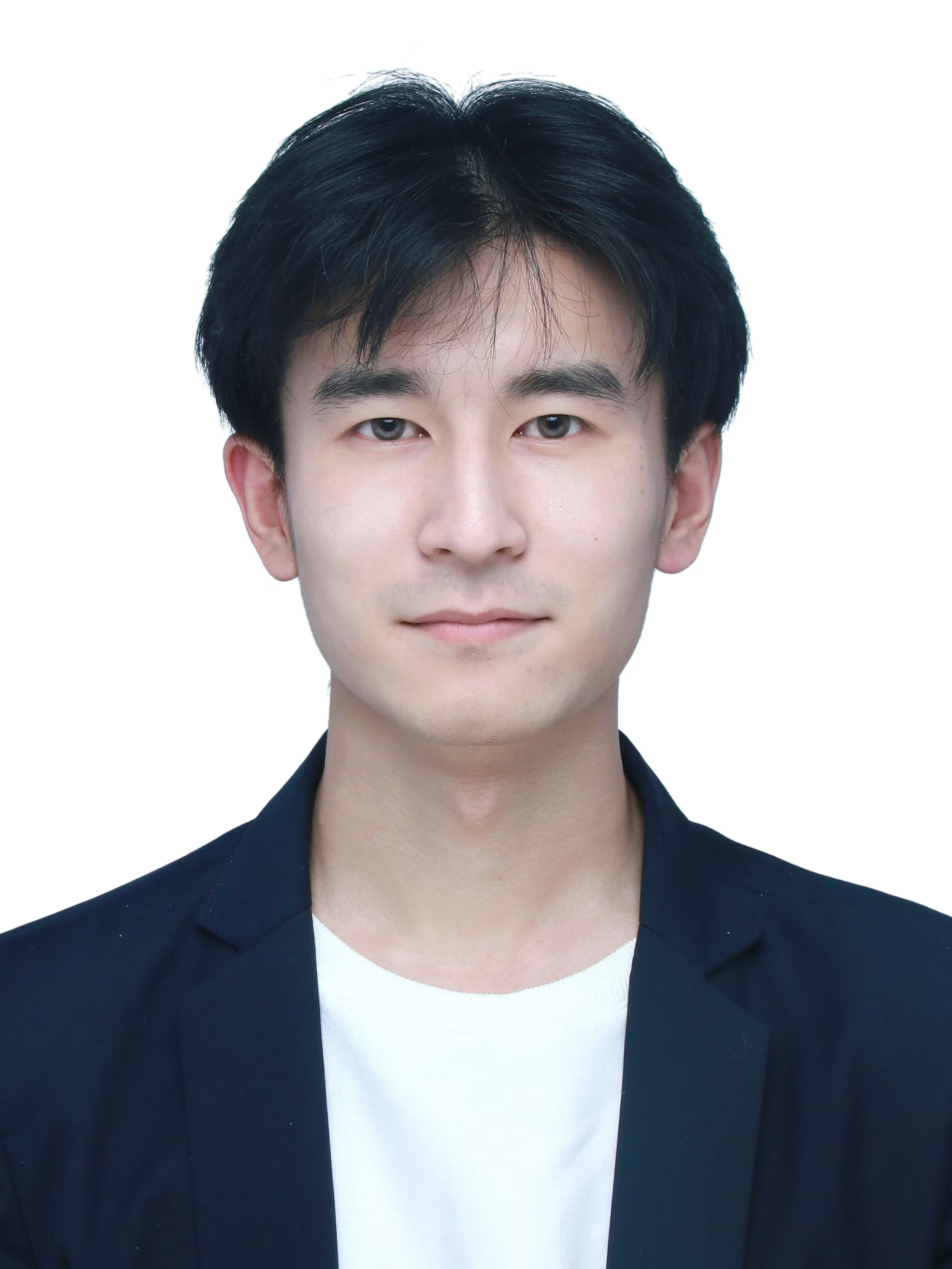}}]{Xiangyu Li} was born in Shandong, China. He received the B.S. degree in software engineering from Dalian University in 2022. Now he is pursuing software engineering in Dalian University and is working hard to pursue a master's degree. Her research interests include deep learning and computer vision.
\end{IEEEbiography}

\vspace{-3mm}
\begin{IEEEbiography}[{\includegraphics[width=1in,height=1.25in,clip,keepaspectratio]{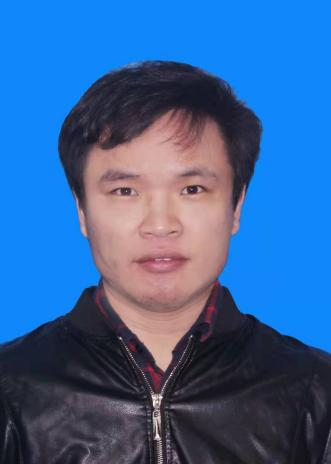}}]{Cong Wang} is currently a Ph.D. student at the Department of Computing, The Hong Kong Polytechnic University. He received the Master’s Degree in Computational
Mathematics from Dalian University of Technology and the Bachelor’s Degree in Mathematics and Applied Mathematics from Inner Mongolia University. His research interests include computer vision and deep learning.
\end{IEEEbiography}

\begin{IEEEbiography} [{\includegraphics[width=1in,height=1.25in,clip,keepaspectratio]{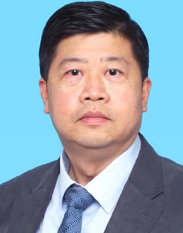}}]{Kin-Man Lam } (Senior Member, IEEE) received the
associateship in electronic engineering with distinc-
tion from the Hong Kong Polytechnic University, in 1986,
the MSc degree in communication engineering from
the Department of Electrical Engineering, Imperial
College, U.K., in 1987, and the PhD degree from
the Department of Electrical Engineering, University
of Sydney, Australia, in 1996. From 1990 to 1993,
he was a lecturer with the Department of Electronic
Engineering, Hong Kong Polytechnic University. He
joined the Department of Electronic and Information Engineering, The Hong
Kong Polytechnic University again as an assistant professor in 1996. He became
an associate professor in 1999, and has been a professor since 2010. Currently,
he is also an associate dean with the Faculty of Engineering. He was actively
involved in professional activities. He was the Chairman of the IEEE Hong
Kong Chapter of Signal Processing between 2006 and 2008, and was the
Director-Student Services and the Director-Membership Services of the IEEE
SPS between 2012 and 2014, and between 2015 and 2017, respectively. He
was also the VP-Member Relations and Development and VP-Publications
of the Asia-Paciﬁc Signal and Information Processing Association (APSIPA)
between 2014 and 2017, and between 2017 and 2021, respectively. He was an
associate editor of IEEE Transaction on Image Processing between 2009 and
2014, and Digital Signal Processing between 2014 and 2018. He was also an
Editor of HKIE Transactions between 2013 and 2018, and an Area Editor of the
IEEE Signal Processing Magazine between 2015 and 2017. Currently, he is the
IEEE SPS VP-Membership and the Member-at-Large of APSIPA. Prof. Lam
also serves as a Senior Editorial Board member of APSIPA Trans. on Signal
and Information Processing and an Associate Editor of EURASIP International
Journal on Image and Video Processing. His current research interests include
image and video processing, computer vision, and human face analysis and
recognition.
\end{IEEEbiography}

\begin{IEEEbiography}[{\includegraphics[width=1in,height=1.25in,clip,keepaspectratio]{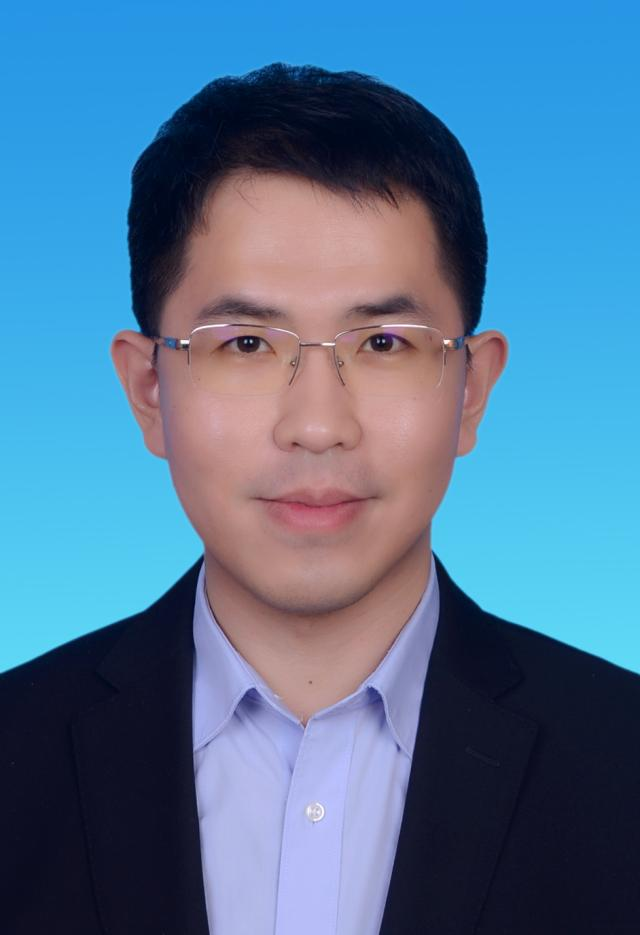}}]{Xin Yang} is a professor and doctoral advisorat Dalian University of Technology. He received his Ph.D. degree in computer science from Zhejiang University (2007-2012), and his B.S. degree in computer science from Jilin University(2003-2007). His main research interests include computer graphics and vision, intelligentrobot technology, focusing on the efficient expression, understanding, perception and interaction of scenes.
\end{IEEEbiography}

\begin{IEEEbiography}[{\includegraphics[width=1in,height=1.25in,clip,keepaspectratio]{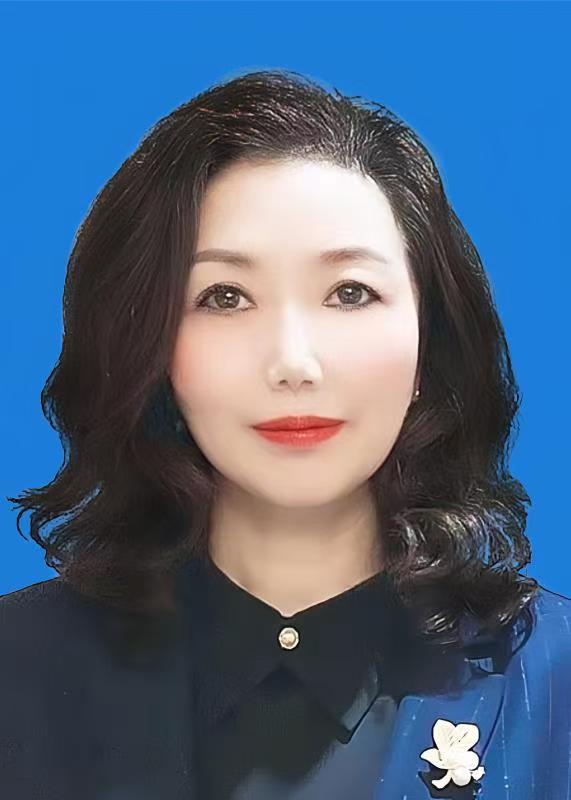}}]{Haiyan Zhang} a professor and master's supervisor at Dalian University, obtained a Master's degree in Financial Asset Management and Finance from Brest Business School, France (2022-2023). Her primary research focuses on the industrialization of brain-computer interface intelligent rehabilitation robots and the industrial application of advanced artificial intelligence technologies in the medical field.
\end{IEEEbiography}

\begin{IEEEbiography}[{\includegraphics[width=1in,height=1.25in,clip,keepaspectratio]{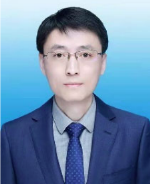}}]{Dongsheng Zhou} (Member, IEEE) received the Ph.D. degree in Dalian University of Technology. Currently, he is a Distinguish Professor of Liaoning Province. In 2019, he established the School of Software Engineering of Dalian University and served as the first dean. His research interests include computer graphics, intelligence computing, and humanrobot interaction. He was appointed as a member of the Computer Science and Technology Discipline Evaluation Group of the Academic Degrees Committee of the Sixth People’s Government of Liaoning Province. He is a member of IEEE, ACM, CGS and CCF.
\end{IEEEbiography}

\end{document}